\documentclass{article}

\usepackage[preprint]{neurips_2026}

\usepackage[utf8]{inputenc} 
\usepackage[T1]{fontenc}    
\usepackage{hyperref}       
\usepackage{url}            
\usepackage{booktabs}       
\usepackage{amsfonts}       
\usepackage{nicefrac}       
\usepackage{microtype}      
\usepackage{xcolor}         
\usepackage{microtype}
\usepackage{graphicx}
\usepackage{subcaption}
\usepackage{multirow}
\usepackage{array}
\usepackage{makecell}
\usepackage{tabularx}   
\usepackage{enumitem}   
\usepackage{longtable}
\usepackage{placeins}
\usepackage{amsmath}
\usepackage{amssymb}
\usepackage{mathtools}
\usepackage{amsthm}
\usepackage{float} 
\usepackage{caption}
\usepackage{xcolor}
\title{Learning Dynamic Belief Graphs for Theory-of-mind Reasoning}

\author{
Ruxiao Chen \\
Department of Civil and Systems Engineering \\
Data Science and AI Institute\\
Johns Hopkins University \\
Baltimore, MD, USA \\
\And
Xilei Zhao \\
Civil and Coastal Engineering \\
University of Florida \\
Gainesville, FL, USA \\
\And
Thomas J. Cova \\
School of the Environment, Society \& Sustainability \\
University of Utah \\
Salt Lake City, UT, USA \\
\And
Frank A. Drews \\
Department of Psychology \\
University of Utah \\
Salt Lake City, UT, USA \\
\And
Susu Xu\thanks{Corresponding author} \\
Department of Civil and Systems Engineering \\
Data Science and AI Institute\\
Johns Hopkins University \\
Baltimore, MD, USA \\
}

\begin{document}

\maketitle

\begin{abstract}
Theory of Mind (ToM) reasoning in highly uncertain environments often involves complex, dynamic, and interdependent belief evolution that jointly shape the process of information seeking and sequential decision-making -- especially in high-stakes settings such as disaster response, emergency medicine, and human-in-the-loop autonomy. 
However, prior approaches for ToM reasoning in LLMs often treat beliefs as static and independent, producing incoherent mental models over time and weak reasoning in dynamic contexts.
We introduce a structured cognitive trajectory model for LLM-based ToM that represents mental state as a dynamic belief graph, jointly inferring latent beliefs, learning their time-varying dependencies, and linking belief evolution to information seeking and decisions.
Our model contributes (i) a novel projection from textualized probabilistic statements to consistent probabilistic graphical model updates, (ii) an energy-based factor graph representation of belief interdependencies, and (iii) an ELBO-based objective that captures belief accumulation and delayed decisions. 
Across multiple real-world disaster evacuation datasets, our model significantly improves action prediction and recovers interpretable belief trajectories consistent with human reasoning, providing a principled module to augment LLMs using ToM reasoning with dynamic and interdependent beliefs in highly uncertain environment.
\end{abstract}

\section{Introduction}
Theory of Mind (ToM) refers to the ability to reason about people’s latent mental states, such as beliefs, intentions, and expectations, and to use these inferences to explain and predict behavior~\cite{tom2009,WIMMER1983103}. 
While ToM has been central in psychology and cognitive science, it is increasingly important for AI systems built on Large Language Models (LLMs) that seek to model, predict, or interact with human behavior~\cite{Baker2017Rational,zhang2025autotom,jin2024mmtom}. 
A key challenge in existing Theory-of-Mind reasoning is that the beliefs underlying human behavior are usually dynamic, interdependent, and unobserved. In real-world decision-making, people rarely act from a single isolated belief. Instead, they maintain multiple beliefs simultaneously, update them as new observations arrive, and allow  different beliefs to reinforce or weaken one another \cite{Christian2024, JARAETTINGER2019105, PADM2012}.

These properties are especially important in high-stakes decision-making, such as emergency response, financial crises, or medical triage~\cite{PADM2012, Croskerry2009,Alfiana2025}, where decisions often unfold through sequential belief updating rather than a one-shot judgment. 
For example, in wildfire evacuation, cognitive theories such as the Protective Action Decision Model (PADM)~\cite{PADM2012} emphasize that protective behavior depends on multiple beliefs, including perceived threats and risks to oneself, property, and others, which jointly shape people's sequential information seeking and evacuation decisions. Similarly, in emergency medicine, Recognition-Primed Decision theories describe how clinicians continuously revise beliefs about symptom severity, disease likelihood, treatment risk, and time pressure as new tests, observations, or patient responses become available before making treatment decisions~\cite{Croskerry2009}. More broadly, Bayesian and predictive-processing accounts characterize cognition as a sequential process of latent belief updating under uncertainty~\cite{JARAETTINGER2019105,Baker2017Rational}. Across these settings, interdependent belief updates are often nonlinear, context-dependent, and rapidly evolving over time~\cite{champ2016,Paton2019DisasterRisk,Cova2024}.

In machine learning, Theory of Mind is typically formalized through Bayesian inverse planning (BIP), which infers mental states via inversion of a generative belief–action process~\cite{Baker2017Rational}.
BIP is principled but typically depends on synthetic state spaces (e.g., grid worlds) and specified dynamics, limiting their scalability to real-world settings. 
Recent LLM-based ToM methods leverage LLMs to infer beliefs or intentions from textual descriptions~\cite{zhang2025autotom,jin2024mmtom,shi2025mumatom} through inference-time prompting and hypothesis sampling to estimate belief distribution. 
However, most existing LLM approaches typically (1) rely on explicit belief signals: they either directly prompt LLMs to output belief states \cite{wilf-etal-2024-think}, or supervise or guide belief inference in LLMs using ground-truth belief labels, which are highly challenging to collect in real-world, high-stakes settings \cite{chen-etal-2025, zhang2025autotom};
(2) reduce belief representations to a small set of independent or mutually exclusive hypotheses~\cite{NEURIPS2023_2b9efb08, le-etal-2019-revisiting,kim-etal-2023-fantom}, which ignores that real mental states involve multiple interdependent beliefs whose interactions can reinforce or suppress one another, so that behavior depends on their joint configuration rather than any single belief;
(3) are formulated in static or single-step settings and lack temporally aligned observation--action trajectories, not capturing how beliefs evolve over time through accumulation and continuous updates as new information arrives.

\begin{figure*}[t]
    \centering
    \includegraphics[width=1\linewidth]{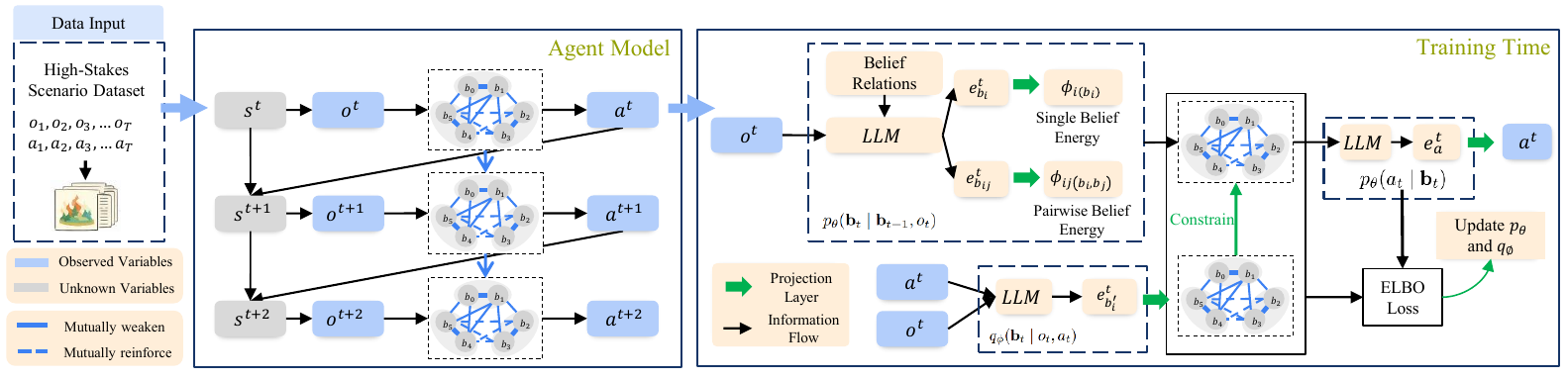}
    \caption{Overview of the Structured Cognitive Trajectory ToM framework. Here, $s_t$ denotes the latent environment state, $o_t$ the agent's observation, $a_t$ the observed action, $b_t$ the latent belief state, and $e_t$ the semantic embedding extracted by a LLM.}
    \vspace{-10pt}
    \label{fig:wide_figure}
\end{figure*}

In this paper, we introduce a structured cognitive trajectory framework for LLM-based ToM reasoning. We represent mental states as dynamic belief graphs, capturing how multiple latent beliefs interact and evolve over time. The framework integrates LLM-derived semantic representations with a structured probabilistic model, enabling belief updates conditioned on observations and past states. Actions are predicted based on the evolving belief state, allowing the model to capture how belief dynamics drive decision-making. The entire system is trained end-to-end using a variational objective that ensures inferred beliefs are behaviorally grounded. At test time, the model performs online belief updating and action prediction from observations alone. An overview of the proposed framework is shown in Fig.~\ref{fig:wide_figure}.
Our contributions include:
\begin{itemize}[topsep=0pt,leftmargin=8pt,itemsep=0pt,parsep=0pt]
\item We introduce a structured cognitive trajectory framework that models mental states as dynamic belief graphs rather than static and independent hypotheses. The framework captures key properties of real-world decisions: multiple co-existing beliefs and sequential belief updating over time.
\item We develop a neuro-symbolic latent variable model that combines LLM-derived semantic evidences with structured probabilistic reasoning. The model jointly learns temporal belief evolution and inter-beleif dependencies through temporally conditioned unary and pairwise belief interactions.
\item We derive a variational learning approach to jointly learn latent belief dynamics directly from observed action trajectories \emph{without any belief-level supervision}. Unlike prior ToM approaches that infer beliefs through prompting or explicit labels, our framework behaviorally grounds latent beliefs through an ELBO objective linking observations, beliefs, and actions.
\item We introduce a stronger evaluation setting for ToM reasoning beyond standard final-answer prediction \footnote{\url{https://anonymous.4open.science/r/Nips_submission-F86B/README.md}}. In addition to action prediction, we evaluate whether inferred latent beliefs align with independently collected human self-reported cognitive states, recover human belief-interaction structures, and exhibit realistic temporal belief dynamics.
\end{itemize}


\section{Related Works}
\noindent\textbf{Machine and LLM-Based Theory of Mind}:
Machine Theory of Mind (ToM) frames mental state inference as a probabilistic inverse problem, where latent beliefs are inferred from observed actions.
A canonical formulation is Bayesian Inverse Planning, which frames ToM reasoning as inverting a generative decision-making model, instantiated as a POMDP~\cite{Baker2017Rational}.
Subsequent work scales this paradigm using neural and amortized inference, while largely retaining manually specified belief structures~\cite{Jha2024}.
While classical Machine ToM provide a general framework for inferring latent mental states, they are typically evaluated in synthetic environments such as grid worlds or small-scale simulated tasks.
Recent work has explored the use of Large Language Models (LLMs) to support mental state inference from textual descriptions and social narratives ~\cite{zhang2025autotom,jin2024mmtom,shi2025mumatom, yang2025, kim2025hypo}.

MuMToM~\cite{jin2024mmtom} integrates LLMs with a predefined POMDP-based agent model for Theory-of-Mind inference, using LLMs to interpret language observations and estimate action likelihoods for inverse planning.
Building on this paradigm, AutoToM~\cite{zhang2025autotom} further removes the need for a fixed agent model by leveraging LLMs to automatically generate belief hypotheses and estimate their posterior probabilities, effectively performing Bayesian-style ToM inference with model structure discovery directly in the language space.
Nevertheless, these methods perform belief inference at inference time using frozen LLM priors, rather than learning belief representations from data.
As a result, beliefs are treated as static and independent hypotheses without statistical grounding or temporal coherence, making them susceptible to hallucinations under unconstrained prompting.

\textbf{Deep Temporal Generative Models.} Deep generative models, particularly Variational Autoencoders (VAEs), provide a principled framework for latent variable inference via variational optimization of the ELBO. To model sequential data, researchers have extended the Variational Autoencoder (VAE) to time-series settings, broadly referred to as Deep Markov Models (DMMs)~\cite{krishnan2016} or Sequential VAEs~\cite{Chung2015}. These approaches model the temporal dynamics of a latent process $z_{1:T}$ via a transition distribution $p_\theta(z_t \mid z_{t-1})$, enforcing temporal coherence across latent states and allowing information to be integrated over time rather than treated independently.

\textbf{Energy-Based Models and Factor Graphs.}
To explicitly model complex dependencies between variables, researchers have utilized Energy-Based Models (EBMs) and Factor Graphs. These approaches define a joint distribution via an energy function that penalizes inconsistent configurations. Recent works like Structured Prediction Energy Networks (SPENs)~\cite{belanger2016} and Factor Graph Neural Networks~\cite{zhang2023} leverage deep learning to parameterize these energy potentials, enabling the learning of higher-order correlations (e.g., in image segmentation or molecular structures).
Our methodology integrates the above two paradigms. We employ a Deep Markov Model to drive the temporal transition of beliefs, while embedding a factor graph at each time step to enforce logical consistency and structured dependencies among the cognitive variables.
\section{Structured Cognitive Trajectory Modeling}
\subsection{Problem Formulation}

Consistent with Theory-of-Mind formulations, we model human decision-making  as a sequential decision-making process. Unlike traditional LLM-based ToM settings that represent an agent's mental state with a single belief or a small set of mutually exclusive beliefs, our setting allows reasoning over multiple co-existing beliefs.  
Specifically, an agent interacts with an environment over discrete time steps $t=1,\dots,T$: at each step, the environment has underlying physical states $s_t$. The agent receives partial observation $o_t$, and then updates their internal latent beliefs $\mathbf{b}_t$, and selects an action $a_t$, which further influences subsequent environmental states. We represent the cognitive state as a structured belief graph $\mathcal{G}_t = (\mathbf{b}_t, \mathcal{E}_t)$, where $\mathbf{b}_t = [b_{t,1}, \dots, b_{t,K}] \in \{0,1\}^K$ 
denotes the set of latent belief variables and $\mathcal{E}_t$ represents time-varying dependencies between beliefs. 

Each belief variable $b_{t,i}$ corresponds to a decision-relevant proposition (e.g., ``my home is in danger'' or ``I might be injured''), while the edges in $\mathcal{E}_t$ capture how beliefs reinforce or suppress one another over time. The model maintains marginal belief probabilities $P(b_{t,i}=1)$, which represent the subjective likelihood assigned to each proposition. Actions such as information seeking, waiting, preparation, or evacuation are generated from the joint belief configuration rather than from any single belief alone, and may further influence future environment states and observations.

The resulting cognitive trajectory is represented as $(s_t, o_t, \mathcal{G}_t, a_t)_{t=1}^T$, where the environment states $s_t$ and belief graphs $\mathcal{G}_t$ are latent, while observations $o_t$ and actions $a_t$ are observed.
Examples of the textual observations and belief semantics associated with each variable are provided in Appendix~\ref{appx:select_q}.

\textbf{Generative Process:}
We model the belief--action process as a structured generative latent-variable model.
Conditioned on the previous belief state and the current observation, the belief transition prior $p_\theta(\mathbf{b}_t \mid \mathbf{b}_{t-1}, o_t)$ captures belief persistence, accumulation, and inter-belief interactions.
Given the latent belief state, the action model $p_\theta(a_t \mid \mathbf{b}_t)$ specifies the distribution over actions.
The joint distribution over belief and action trajectories factorizes as:
{\setlength{\abovedisplayskip}{4pt}
 \setlength{\belowdisplayskip}{4pt}
\begin{equation}
    p(a_{1:T}, \mathbf{b}_{1:T} \mid o_{1:T}) 
    = \prod_{t=1}^{T} 
    \underbrace{p_\theta(a_t \mid \mathbf{b}_t)}_{\text{Action Model}}
    \cdot
    \underbrace{p_\theta(\mathbf{b}_t \mid \mathbf{b}_{t-1}, o_t)}_{\text{Belief Transition Prior}}.
\end{equation}}

At each time step, the belief state $\mathbf{b}_t$ is modeled as a Markov Random Field whose dependencies evolve over time. To instantiate this structured latent belief framework, we next introduce a neuro-symbolic generative model that converts LLM-derived semantic evidence into coherent probabilistic belief dynamics and belief-conditioned decision-making.

\subsection{Neuro-Symbolic Generative Model}
Though LLMs provide rich semantic representations, independent outputs alone do not define a coherent joint belief distribution: they may produce temporally inconsistent beliefs \cite{steyvers2025llmknowledge}, ignore inter-belief dependencies, and lack behavioral grounding \cite{sclar-etal-2023-minding}.  To address this, we use LLM-derived semantic representations as contextual evidence rather than directly treating them as probabilities of beliefs. These semantic representations are projected into a structured probabilistic latent-variable model that defines coherent unary and pairwise interactions among beliefs, enabling globally consistent temporal belief evolution and belief-conditioned decision-making under uncertainty. In this section, we introduce a neuro-symbolic generative model that combines LLM-derived semantic representations with structured probabilistic reasoning over latent belief trajectories.

\noindent\textbf{Semantic-to-Potential Projection:}
Conditioned on the previous belief state $\mathbf{b}_{t-1}$ and the current observation $o_t$, we define the belief transition prior as a Gibbs distribution \cite{LeCun2006ATO} over the joint configuration of $K$ binary beliefs.
And the corresponding energy function is:
{\setlength{\abovedisplayskip}{4pt}
 \setlength{\belowdisplayskip}{4pt}
\begin{equation}
E_\theta(\mathbf{b}_t)
= -\sum_{i=1}^{K} \phi_i(b_{t,i})
- \sum_{1 \le i < j \le K} \phi_{ij}(b_{t,i}, b_{t,j}),
\end{equation}}

where $\phi_i(b_{t,i})$ and $\phi_{ij}(b_{t,i}, b_{t,j})$ denote the unary and pairwise potentials, respectively.
These potentials are parameterized by LLM-based semantic evidence and are implicitly conditioned on $(\mathbf{b}_{t-1}, o_t)$.
The belief transition prior is defined as $p_\theta(\mathbf{b}_t \mid \mathbf{b}_{t-1}, o_t) = \frac{\exp\!\big(-E_\theta(\mathbf{b}_t)\big)}{Z_q(\mathbf{b}_{t-1}, o_t)}$, where $Z_q(\mathbf{b}_{t-1}, o_t)$ is the partition function that normalizes the distribution over all $2^K$ belief configurations.

We next describe how LLM embeddings are used to parameterize the unary and pairwise potentials.
For each belief $i$, we query a frozen LLM to obtain semantic embedding representations that assess whether the resident is likely to hold the target belief under current observation $o_t$, while accounting for belief persistence over time.
Specifically, we construct two prompt variants that differ only in the assumed previous belief state: one with $b_{t-1,i}=1$ (``Yes'') and one with $b_{t-1,i}=0$ (``No''), yielding $\mathbf{h}^{\text{Yes}}_{t,i} = \mathrm{LLM}(b_{t,i}\mid o_t,\, b_{t-1,i}{=}1)$ and $\mathbf{h}^{\text{No}}_{t,i} = \mathrm{LLM}(b_{t,i}\mid o_t,\, b_{t-1,i}{=}0)$.
To incorporate belief history into current evidence extraction in a differentiable and semantically consistent manner, we take the expectation of the LLM embedding over the previous belief state under the belief transition model:
\begin{equation}
\mathbf{h}_{t,i}
=
p_\theta(b_{t-1,i}=1)\, \mathbf{h}^{\text{Yes}}_{t,i}
+
\big(1 - p_\theta(b_{t-1,i}=1)\big)\, \mathbf{h}^{\text{No}}_{t,i}.
\end{equation}
where $p_\theta(b_{t-1,i}=1)$ denotes the model-implied marginal belief probability from prior timestep $t\!-\!1$.

\emph{\underline{Unary Potential and Pairwise Potential Function}}: Because belief states are unobserved, unary potentials are learned in an unsupervised setting where the sign of the energy is not identifiable, as flipping the semantic interpretation of a belief yields an equivalent likelihood.
To prevent such semantic flipping, we anchor the unary energy to the LLM’s semantic direction by contrasting its alignment with the two reference embeddings. We define this base unary score as $\phi^{\text{base}}_{t,i}
= \tau\!\left(
\cos(\mathbf{h}_{t,i}, \mathbf{h}^{\text{Yes}}_{t,i})
- \cos(\mathbf{h}_{t,i}, \mathbf{h}^{\text{No}}_{t,i})
\right)$, where $\tau$ is a temperature parameter.
This construction enforces a consistent semantic orientation of the latent belief space.
Finally, we learn a residual value on top of this anchored signal to capture deviations not explained by the contrastive alignment:
\begin{equation}
\phi_{t,i}
= \phi^{\text{base}}_{t,i}
+ \mathbf{w}_u^\top \mathrm{ReLU}(\mathbf{h}_{t,i})
+ \beta_u .
\end{equation}
We further design a pairwise embedding for each belief pair as:
\begin{equation}
\phi_{ij} = \mathbf{w}_p^\top \mathrm{ReLU}(\mathbf{h}_{ij}) + \beta_p .
\end{equation} 
where $\mathbf{h}_{ij} = \mathrm{LLM}(b_i,b_j)$ is derived from LLM embedding.
Intuitively, $\phi_{ij}>1$ encourages co-activation of beliefs $i$ and $j$, while $0<\phi_{ij}<1$ discourages co-activation.

\emph{\underline{Marginal Belief Probabilities}}: Given the belief transition prior $p_\theta(\mathbf{b}_t \mid \mathbf{b}_{t-1}, o_t)$ defined in Eq.~(2), we compute marginal belief probabilities by summing over all joint belief configurations.
Let $\mathbf{b}_t = (b_{t,1}, \dots, b_{t,K}) \in \{0,1\}^K$ denote a joint belief assignment.
The marginal probability that belief $i$ is active at time $t$ is given by $p_\theta(b_{t,i}=1 \mid \mathbf{b}_{t-1}, o_t) = \sum_{\mathbf{b}_t:\, b_{t,i}=1} p_\theta(\mathbf{b}_t \mid \mathbf{b}_{t-1}, o_t)$.

\emph{\underline{Action Generation}}: Given belief marginals, we construct belief-conditioned embeddings for each action $j \in \mathcal{A}$.
For each belief $i$, we precompute two action embeddings corresponding to $b_i = 1$ and $b_i = 0$, which are represented as:
$\mathbf{e}^{(1)}_{j,i}=\text{LLM}(a_j \mid b_i{=}1)$
and
$\mathbf{e}^{(0)}_{j,i}=\text{LLM}(a_j \mid b_i{=}0)$.
and mix them using the marginal belief probability
{\setlength{\abovedisplayskip}{4pt}
 \setlength{\belowdisplayskip}{4pt}
\begin{equation}
\mathbf{x}_{t,j,i}
=
p_\theta(b_{t,i}=1)\, \mathbf{e}^{(1)}_{j,i}
+
\big(1-p_\theta(b_{t,i}=1)\big)\, \mathbf{e}^{(0)}_{j,i},
\qquad
\mathbf{A}_{t,j}
=
\mathrm{softmax}\!\left(
\frac{\mathbf{Q}_{t,j}\mathbf{K}_{t,j}^\top}{\sqrt{d_k}}
\right).
\end{equation}}

Stacking $\{\mathbf{x}_{t,j,i}\}_{i=1}^K$ yields an action-specific belief token matrix $\mathbf{X}_{t,j}\in\mathbb{R}^{K\times d}$. Because the influence of beliefs on actions is inherently non-linear—different \emph{combinations} of beliefs can trigger different actions, and one belief may modulate the relevance of another—we apply self-attention over belief-conditioned tokens separately for each action.
For a given action $j$, the input to the attention module is the belief token matrix $\mathbf{X}_{t,j}\in\mathbb{R}^{K\times d}$, where each row corresponds to one belief dimension.
Self-attention maps $\mathbf{X}_{t,j}$ to an interaction-aware representation $\mathbf{Z}_{t,j}\in\mathbb{R}^{K\times d}$ by computing an action-specific attention matrix \cite{vaswani2023}
where $\mathbf{Q}_{t,j}, \mathbf{K}_{t,j}, \mathbf{V}_{t,j}$ are linear projections of $\mathbf{X}_{t,j}$.
The entry $\mathbf{A}_{t,j}[i,k]$ quantifies how belief $k$ modulates the contribution of belief $i$ when forming the decision representation for action $j$.
The resulting interaction-aware belief features are then aggregated into an action-level representation, which parameterizes the action likelihood.
Since the true environment state is unobserved, after sampling an action we query the LLM to generate a textual pseudo-state $s_{t+1}$, whose description is used as the next observation $o_{t+1}$, consistent with Theory-of-Mind assumptions on latent state transitions.

\subsection{Inference Model}
In realistic settings, latent beliefs are typically not directly observable, and ground truth of evolving beliefs is often unavailable. Here we formulate belief inference as an unsupervised latent-variable problem: given observations $o_t$ and observed actions $a_t$, the model must infer the latent belief configuration that plausibly explains the behavior. Formally, this corresponds to inference over the posterior  $p_\theta(\mathbf{b}_t \mid o_t, a_t)$, which is intractable due to the structured temporal dependencies and inter-belief interactions in the generative model. To enable efficient inference, we introduce an amortized variational posterior $q_\phi(\mathbf{b}_t \mid o_t, a_t)$ to approximate the intractable posterior over beliefs during learning. Because the observed action $a_t$ provides additional evidence about the latent beliefs that likely generated the behavior, the inference model conditions on both observations and realized actions. 
We parameterize the variational posterior in a factorized form:
{\setlength{\abovedisplayskip}{4pt}
 \setlength{\belowdisplayskip}{4pt}
\begin{equation}
q_\phi(\mathbf{b}_t \mid o_t, a_t)
=
\prod_{i=1}^{K}
q_\phi(b_{t,i} \mid \mathbf{u}^{\mathrm{inf}}_{t,i}), \quad
\mathbf{u}^{\mathrm{inf}}_{t,i}
=
\mathrm{LLM}(b_{t,i} \mid o_t, a_t).
\end{equation}}

where $\mathbf{u}^{\mathrm{inf}}_{t,i}$ is a belief-specific semantic embedding extracted by a frozen LLM.
Conditioning on both $o_t$ and $a_t$ allows the inference model to attribute observed actions to plausible underlying beliefs.
Each belief posterior is modeled as a Bernoulli distribution with logit: $\ell_{t,i} = \mathbf{W}_1 \,\mathrm{ReLU}\!\left(\mathbf{u}^{\mathrm{inf}}_{t,i}\right) + \mathbf{b}_1$.
Here, $i\in\{1,\dots,K\}$ indexes the belief dimension, and $\ell_{t,i}\in\mathbb{R}$ denotes the Bernoulli logit for $b_{t,i}$.
\subsection{Variational Learning Objective}
Since beliefs $\mathbf{b}_t$ are latent and not directly observed, we train the model by maximizing an evidence lower bound (ELBO) \cite{kingma2022, Rezende2014} on the conditional log-likelihood of actions.
At each time step $t$, the generative model defines the joint distribution $p_\theta(a_t, \mathbf{b}_t \mid \mathbf{b}_{t-1}, o_t) = p_\theta(a_t \mid \mathbf{b}_t)\, p_\theta(\mathbf{b}_t \mid \mathbf{b}_{t-1}, o_t)$.
where $p_\theta(\mathbf{b}_t \mid \mathbf{b}_{t-1}, o_t)$ is the temporal belief transition prior (Eq.~7) and $p_\theta(a_t \mid \mathbf{b}_t)$ is the belief-conditioned action likelihood parameterized by the attention-based action model (Eq.~9).
Because the posterior $p_\theta(\mathbf{b}_t \mid o_t, a_t)$ does not admit a
tractable form for stable end-to-end optimization and amortized inference,
we approximate it with the inference model $q_\phi(\mathbf{b}_t \mid o_t, a_t)$.
Applying standard variational inference, the training objective is defined as a sum of per-time-step ELBOs:
{\setlength{\abovedisplayskip}{4pt}
 \setlength{\belowdisplayskip}{4pt}
\begin{equation}
\mathcal{L}
= -\sum_{t=1}^{T}
\big(\mathcal{L}^{\text{act}}_t - \mathcal{L}^{\text{KL}}_t\big).
\end{equation}}

The action likelihood term encourages belief configurations that explain the observed action, and the KL term enforces consistency between the inference posterior and the belief transition prior:
{\setlength{\abovedisplayskip}{4pt}
 \setlength{\belowdisplayskip}{4pt}
\begin{equation}
\mathcal{L}^{\text{act}}_t
=
\mathbb{E}_{q_\phi(\mathbf{b}_t \mid o_t, a_t)}
\big[
\log p_\theta(a_t \mid \mathbf{b}_t)
\big], \quad
\mathcal{L}^{\text{KL}}_t
=
\mathrm{KL}\!\left(
q_\phi(\mathbf{b}_t \mid o_t, a_t)
\,\Vert\,
p_\theta(\mathbf{b}_t \mid \mathbf{b}_{t-1}, o_t)
\right).
\end{equation}}

As mentioned earlier, the inference model conditions on the realized action $a_t$, whereas the generative prior does not.
This standard variational asymmetry allows action information to guide posterior belief inference during training and be transferred to the generative model through the KL term.
A detailed derivation of the ELBO is provided in Appendix~\ref{appx:elbo}.
At test time, only the generative components $p_\theta(\mathbf{b}_t \mid \mathbf{b}_{t-1}, o_t)$ and $p_\theta(a_t \mid \mathbf{b}_t)$ are used.

\begin{table*}[t]
\centering
\footnotesize
\setlength{\tabcolsep}{3pt}
\renewcommand{\arraystretch}{0.95}

\caption{
Action prediction performance on wildfire datasets with sequential multi-belief updates.
All metrics are higher-is-better ($\uparrow$); the best result in each row is shown in \textbf{bold}.
$\ddagger$ and $\dagger$ denote multi-step and single-step reasoning, respectively; for benchmarks, they indicate multi-step or single-step tasks.
$\star$ denotes unsupervised latent belief learning.
$\circ$ denotes methods that rely on supervised belief learning or external feedback during belief inference. LLM Inference and SimToM are direct inference methods; SimToM uses a two-stage prompting framework.
} 

\label{tab:all_llm_benchmark_action_results}

\resizebox{\textwidth}{!}{

\begin{tabular}{lll cc cc cc cc cc}

\toprule

\multirow{2}{*}{Benchmark}
& \multirow{2}{*}{Backbone}
& \multirow{2}{*}{Metric}
& \multicolumn{2}{c}{\textbf{Ours}$^{\star\ddagger}$}
& \multicolumn{2}{c}{AutoToM$^{\circ\ddagger}$}
& \multicolumn{2}{c}{LLM Inference$^{\dagger}$}
& \multicolumn{2}{c}{FLARE$^{\circ\dagger}$}
& \multicolumn{2}{c}{Reconciliation$^{\circ\dagger}$} \\

\cmidrule(lr){4-5}
\cmidrule(lr){6-7}
\cmidrule(lr){8-9}
\cmidrule(lr){10-11}
\cmidrule(lr){12-13}

& &
& Inter & Final
& Inter & Final
& Inter & Final
& Inter & Final
& Inter & Final \\

\midrule

\multirow{12}{*}{\makecell{Wildfire$^\ddagger$ \\ (Six Beliefs)}}

& \multirow{2}{*}{Qwen3-8B}
& Action Acc.
& \textbf{0.8443} & \textbf{0.9157}
& 0.3860 & 0.6959
& 0.2796 & 0.6888
& 0.6790 & 0.7760
& 0.2427 & 0.6194 \\

&
& Macro-F1
& \textbf{0.7960} & \textbf{0.9023}
& 0.1825 & 0.5215
& 0.1629 & 0.4079
& 0.4104 & 0.7741
& 0.1522 & 0.6190 \\

\cmidrule(lr){2-13}

& \multirow{2}{*}{Qwen3-32B}
& Action Acc.
& \textbf{0.9201} & \textbf{0.9490}
& 0.3833 & 0.7742
& 0.3096 & 0.6888
& 0.6607 & 0.7763
& 0.4568 & 0.7373 \\

&
& Macro-F1
& \textbf{0.9310} & \textbf{0.9368}
& 0.1390 & 0.5330
& 0.1182 & 0.4079
& 0.4492 & 0.6556
& 0.1571 & 0.6508 \\

\cmidrule(lr){2-13}

& \multirow{2}{*}{Qwen3-72B}
& Action Acc.
& \textbf{0.8208} & \textbf{0.9005}
& 0.1499 & 0.7788
& 0.3277 & 0.6888
& 0.4821 & 0.5520
& 0.4741 & 0.8341 \\

&
& Macro-F1
& \textbf{0.7855} & \textbf{0.9000}
& 0.0790 & 0.5600
& 0.1766 & 0.4079
& 0.3409 & 0.5448
& 0.1875 & 0.7187 \\

\cmidrule(lr){2-13}

& \multirow{2}{*}{Llama-3-8B}
& Action Acc.
& \textbf{0.7477} & \textbf{0.8677}
& 0.3660 & 0.8387
& 0.3156 & 0.6888
& 0.5714 & 0.7230
& 0.6523 & 0.8111 \\

&
& Macro-F1
& \textbf{0.5842} & \textbf{0.8392}
& 0.2750 & 0.4560
& 0.1433 & 0.4079
& 0.4620 & 0.6249
& 0.3458 & 0.6193 \\

\cmidrule(lr){2-13}

& \multirow{2}{*}{Llama-3-70B}
& Action Acc.
& \textbf{0.9367} & \textbf{0.9286}
& 0.3746 & 0.8387
& 0.3096 & 0.6888
& 0.6607 & 0.7760
& 0.4569 & 0.8571 \\

&
& Macro-F1
& \textbf{0.9456} & \textbf{0.9151}
& 0.1530 & 0.4560
& 0.1182 & 0.4079
& 0.4488 & 0.6964
& 0.1571 & 0.5999 \\

\cmidrule(lr){2-13}

& \multirow{2}{*}{Deepseek-R1 70B}
& Action Acc.
& \textbf{0.9084} & \textbf{0.9592}
& 0.3862 & 0.8433
& 0.1353 & 0.6167
& 0.6250 & 0.6447
& 0.5747 & 0.8710 \\

&
& Macro-F1
& \textbf{0.9133} & \textbf{0.9500}
& 0.1390 & 0.4580
& 0.2287 & 0.4204
& 0.4221 & 0.6016
& 0.3170 & 0.6721 \\
\bottomrule

\end{tabular}
}
\vspace{-10pt}
\end{table*}

\section{Results}
\subsection{Experimental Setup}

\begin{figure*}[t]
    \centering
    \includegraphics[width=1\linewidth]{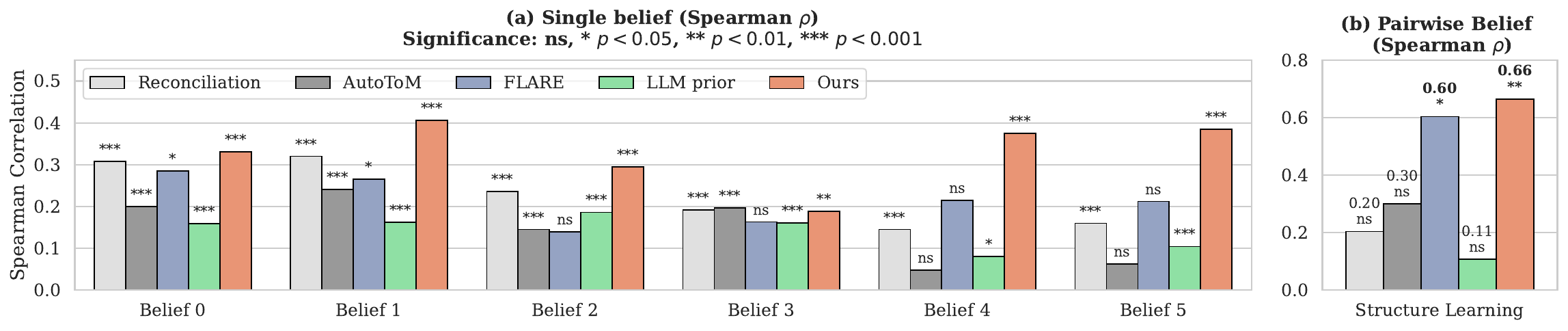}
    \caption{(a) Spearman correlation between model-predicted beliefs and human ratings for individual beliefs. (b) Spearman correlation for pairwise belief structure learning. }
    \vspace{-15pt}
    \label{fig:belief_corr}
\end{figure*}

\underline{\emph{Dataset:}}
Existing ToM benchmarks provide controlled evaluation settings but are typically single-step and only contain one to two types of beliefs in each sample (e.g, BigToM~\cite{NEURIPS2023_2b9efb08}, SimpleToM~\cite{le-etal-2019-revisiting}, and OpenToM~\cite{xu2024opentomcomprehen}). 
To address this gap, we leverage a real-world wildfire evacuation dataset~\cite{KULIGOWSKI2022105541, FORRISTER2024100729}, which contains multi-step observation--action trajectories collected from participants affected by major wildfire events (e.g., Kincade and Marshall fires). The dataset records updates of individuals’ six types of belief over time through structured survey responses (e.g., Likert-scale ratings of perceived risks) and free-text descriptions of their evolving perceptions, information processing, and decision-making. It includes multiple co-existing belief variables, enabling analysis of inter-belief interactions and temporal belief evolution. The exact prompts used to extract belief- and action-conditioned hidden embeddings from the LLM are detailed in Appendix~\ref{appx:prompts}.

We also evaluate on standard ToM benchmarks for comparison (e.g, BigToM, SimpleToM, and OpenToM mentioned above), while noting that these datasets involve limited belief variables and lack temporal structure.

\underline{\emph{Baselines:}}
We compare against these representative baselines.
\emph{AutoToM}~\cite{zhang2025autotom} performs Bayesian-style ToM inference using LLM-generated hypotheses over latent mental variables.
\emph{Model Reconciliation (LLM-based)}~\cite{tang2025model} uses LLMs to generate post-hoc causal explanations when model predictions disagree with observed human decisions.
\emph{FLARE}~\cite{chen-etal-2025} integrates PADM with LLM reasoning to model latent risk perceptions and predict evacuation behavior.
\emph{SimToM}~\cite{wilf-etal-2024-think} performs two-stage prompting with explicit perspective-taking before answering ToM questions.
For the wildfire dataset, we compare against AutoToM, Model Reconciliation, and FLARE. For general ToM benchmarks, we compare against AutoToM and SimToM, as these methods are specifically proposed for ToM reasoning tasks. We also include an LLM prior baseline (direct LLM inference). LLM Inference and SimToM are direct inference baselines without belief learning. FLARE relies on supervised belief signals, while AutoToM and Reconciliation use iterative feedback during inference. In contrast, our model learns latent beliefs without supervision. Both our model and AutoToM model multi-step belief dynamics, whereas the remaining baselines are single-step.

\subsection{Results on Dynamic Multi-belief Wildfire Dataset}
\underline{\emph{ Action Prediction Performance:}}
For wildfire dataset, we evaluate two tasks: intermediate action prediction (4-way classification) and final evacuation prediction (binary).
Accuracy alone is insufficient since biased models may over-predict dominant classes. We therefore report macro-F1 to assess prediction balance. Plain LLM shows strong bias, with very low macro-F1 despite reasonable accuracy, indicating highly imbalanced predictions.
As shown in Table~\ref{tab:all_llm_benchmark_action_results}, our model improves both accuracy and macro-F1, especially on the intermediate task, indicating better capture of diverse decision patterns. These gains suggest that the learned belief graph supports both coarse-grained final decisions and fine-grained intermediate actions. Additional results on ELBO convergence and training dynamics are provided in Appendix~\ref{appx:more_results}. 

\underline{\emph{ Unary Belief Correlation Results:}}
A central goal of our framework is to infer latent cognitive beliefs that mediate decision-making.
Although beliefs are unobserved during training, the post-event wildfire surveys include self-reported belief assessments that can be used as proxy ground truth for evaluation.
This enables validation of inferred beliefs without violating the model’s latent-variable assumption.
The survey measures six belief dimensions relevant to wildfire evacuation, with respondents rating each belief on a 1--5 ordinal scale.
Because these ratings are subjective and not directly comparable across individuals, we evaluate belief quality using Spearman correlation between predicted belief scores and survey responses.
Spearman correlation is a rank-based measure that captures monotonic agreement and is invariant to scale transformations, making it a standard choice for comparing latent belief predictions against ordinal human ratings \cite{Nan2024, Keles2021}.
Figure~\ref{fig:belief_corr} (a) reports per-belief Spearman correlations for our method and all baselines.
Across most belief dimensions, our model achieves substantially higher correlations than all baseline models. Results for beliefs on additional backbones are in Appendix~\ref{appx:allmodels},

\begin{figure*}[t]
    \centering
    \includegraphics[width=1\linewidth]{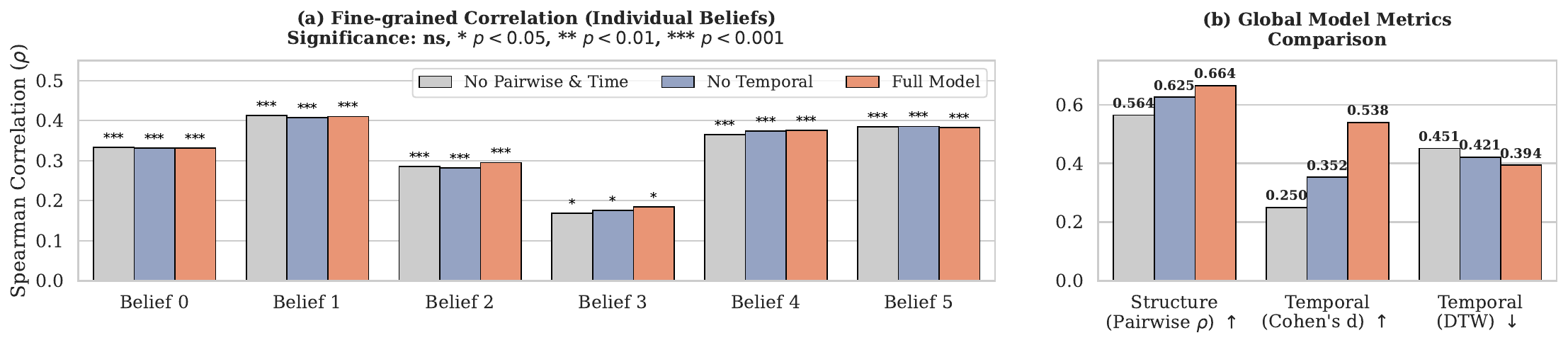}
    \caption{Ablation results: (a) Per-belief Spearman correlation under different ablations. (b) Global metrics for belief structure learning and temporal consistency.}
    \vspace{-15pt}
    \label{fig:ablation}
\end{figure*}

\underline{\emph{ Pairwise Belief Correlation Results:}}We evaluate whether the model captures belief--belief interactions by testing if co-variation patterns among beliefs are preserved: i.e., when belief $i$ tends to increase together with belief $j$ in the survey data, the same tendency should also appear in the model predictions.
For each belief pair $(i,j)$ with $i<j$, we compute Spearman correlations $r^{\mathrm{gt}}_{ij}$ and $r^{\mathrm{pred}}_{ij}$ over individuals using ground-truth and predicted beliefs, respectively.
Agreement between belief structures is then measured by the Spearman correlation, which evaluates whether belief pairs that co-vary more strongly in human data are ranked similarly by the model.
Figure~\ref{fig:belief_corr} (b) shows that our model best recovers the belief–belief co-variation structure, while the other baselines exhibit weaker agreement with the survey-derived interaction ranking.

\underline{\emph{Belief Scalability Analysis}}
Our model performs belief inference via exact enumeration over the joint belief space, which grows exponentially with the number of beliefs. We further consider mean-field and MCMC as approximation strategies and compare them against exact enumeration to evaluate their efficiency and performance trade-offs. Results show that exact inference remains practical up to $K{=}15$, while mean-field provides a more efficient alternative as the belief space grows. Detailed results are provided in Appendix~\ref{appx:scalability}.

\subsection{Results on Other ToM Benchmarks} 

We compare both belief and action prediction on standard text-based ToM benchmarks (Table~\ref{tab:tom_combined}). Our model infers beliefs as latent variables without any belief supervision, whereas other baselines rely on belief labels for supervised learning or feedback during inference. As a result, our belief prediction is often among the highest-performing methods, although not consistently the top across all settings. Importantly, our model consistently achieves strong action prediction performance, often matching or outperforming baselines, showing that accurate decisions can be obtained from latent, behaviorally grounded beliefs without direct belief supervision. Notably, these benchmarks are single-step and involve only a small number of mutually exclusive belief variables, which largely reduce the need for modeling belief dynamics and inter-belief interactions. As a result, our framework performs competitively, but its advantages are less pronounced in this simplified setting.

For additional results, we further analyze the role of latent beliefs through belief mediation (Appendix~\ref{appx:beliefs_mediating}), showing that manipulating inferred beliefs leads to systematic changes in action predictions, confirming causal influence of beliefs. Attention analysis (Appendix~\ref{appx:more_results}) further interprets the action-specific self-attention matrix, showing how one belief dimension modulates another belief's contribution under different target actions.
In addition, the model remains robust under different prompt settings (Appendix~\ref{appx:prompts_sen}), with only minor performance variations. Appendix~\ref{appx:scalability} provides a belief-number ablation and scalability test, analyzing how different belief dimensions affect performance and how exact versus approximate inference scales with larger belief spaces. Sensitivity analysis over $\tau$ is discussed in Appendix ~\ref{appx:more_results}.

\subsection{Ablation Study}
We conduct ablation studies by selectively removing key elements of the proposed model.
In the first ablation (\textsc{No-Pairwise}), we remove the pairwise belief interaction terms $\phi_{ij}$ from the belief transition prior, reducing the temporal MRF to independent unary potentials.
In the second ablation (\textsc{No-Temporal}), we remove temporal belief transitions by discarding dependence on $\mathbf{b}_{t-1}$ and training the model independently at each time step.
All other components are kept unchanged.
Figure~\ref{fig:ablation}(b) compares model variants in terms of how well they recover belief--belief co-variation patterns.
The full model achieves the highest Spearman correlation, while removing pairwise potentials or temporal transitions substantially degrades performance.

We evaluate belief dynamics using two complementary metrics.
The Standardized Mean Difference (Cohen’s d) \cite{Lakens2013, David2019} assesses whether belief updates are preferentially aligned with action changes, capturing whether temporal belief evolution is decision-relevant rather than uniformly drifting.
Dynamic time warping (DTW) distance \cite{Zhou2013, PASTEUNING2025119833} evaluates whether the overall temporal shape of predicted belief trajectories matches human-reported belief evolution, reflecting global temporal consistency.
Together, these metrics distinguish action-aligned belief updates from mere temporal smoothness; implementation details are provided in Appendix~\ref{appx:metrics}.
As shown in Figure~\ref{fig:ablation}(b), the full model achieves the highest Cohen’s d and the lowest DTW distance among all variants.
Figure~\ref{fig:ablation} (a) shows per-belief Spearman correlations across ablations. We observe that even without pairwise or temporal components,  the model achieves comparable performance at the single-belief level, indicating that the ELBO-based latent-variable framework alone suffices to recover marginal belief signals. 
Further ablations on the ELBO objective and factor graph structure, disentangling their contributions to belief learning and action prediction, are provided in Appendix~\ref{appx:beliefs_mediating}.

\begin{table*}[t]
\centering
\scriptsize
\setlength{\tabcolsep}{3pt}
\renewcommand{\arraystretch}{0.95}
\caption{Comparison across Our Model, AutoToM, SimToM, and LLM Inference on SimpleToM, BigToM, and OpenToM. $\ddagger$ and $\dagger$ denote multi-step and single-step reasoning, respectively (or tasks for benchmarks),
$\star$ denotes unsupervised latent belief learning.
$\circ$ denotes methods that rely on supervised belief inference or external feedback during belief inference.
}

\begin{tabular}{llcccccccc}
\toprule
\multirow{2}{*}{Benchmark} & \multirow{2}{*}{Backbone}
& \multicolumn{2}{c}{Our Model $^{\star}$$^{\ddagger}$}
& \multicolumn{2}{c}{AutoToM $^{\circ}$$^{\ddagger}$}
& \multicolumn{2}{c}{SimToM $^{\dagger}$}
& \multicolumn{2}{c}{LLM Inference $^{\dagger}$} \\
\cmidrule(lr){3-4}\cmidrule(lr){5-6}\cmidrule(lr){7-8}\cmidrule(lr){9-10}
& & Belief Acc. & Action Acc.
& Belief Acc. & Action Acc.
& Belief Acc. & Action Acc.
& Belief Acc. & Action Acc. \\
\midrule

\multirow{5}{*}{\makecell{SimpleToM$^\ddagger$ \\ (Single Belief)}}
& DeepSeek-R1    & 0.917 & \textbf{0.996} & \textbf{0.942} & 0.775 & 0.610 & 0.580 & 0.860 & 0.732 \\
& Llama-3-8B     & \textbf{0.987} & \textbf{0.996} & 0.905 & 0.661 & 0.620 & 0.900 & 0.689 & 0.505 \\
& Llama-3-70B    & 0.857 & \textbf{0.933} & 0.979 & 0.847 & \textbf{0.980} & 0.930 & 0.968 & 0.756 \\
& Qwen3-32B      & 0.948 & \textbf{0.991} & 0.906 & 0.746 & \textbf{0.980} & 0.950 & 0.947 & 0.767 \\
& Qwen3-8B       & \textbf{0.961} & \textbf{0.991} & 0.867 & 0.446 & 0.870 & 0.780 & 0.823 & 0.518 \\
\midrule

\multirow{5}{*}{\makecell{BigToM$^\ddagger$ \\ (Two Beliefs)}}

& Qwen3-8B       & \textbf{0.949} & \textbf{0.880} & 0.695 & 0.700 & 0.615 & 0.780 & 0.830 & 0.675 \\
& Llama-3-70B    & 0.966 & 0.870 & \textbf{0.978} & 0.870 & 0.530 & 0.540 & 0.812 & \textbf{0.880} \\
& DeepSeek-R1    & \textbf{0.983} & \textbf{0.870} & 0.975 & 0.863 & 0.740 & 0.830 & 0.890 & 0.790 \\
& Qwen2.5-72B    & 0.949 & \textbf{0.950} & \textbf{0.985} & 0.885 & 0.765 & 0.820 & 0.950 & 0.890 \\
& Qwen3-32B      & 0.875 & \textbf{0.890} & \textbf{0.900} & 0.870 & 0.355 & 0.740 & \textbf{0.900} & 0.870 \\
\midrule

\multirow{5}{*}{\makecell{OpenToM$^\ddagger$ \\ (Two Beliefs)}}

& DeepSeek-R1    & \textbf{0.985} & \textbf{0.940} & 0.730 & 0.680 & 0.850 & 0.760 & 0.760 & 0.860 \\
& Llama-3-8B     & 0.940 & \textbf{0.970} & 0.720 & 0.670 & \textbf{0.970} & 0.930 & 0.930 & \textbf{0.970} \\
& Llama-3-70B    & \textbf{0.945} & 0.895 & 0.800 & 0.820 & 0.860 & 0.620 & 0.860 & \textbf{0.910} \\
& Qwen3-32B      & \textbf{0.910} & 0.885 & 0.740 & 0.850 & 0.820 & 0.830 & 0.810 & \textbf{0.910} \\
& Qwen3-8B       & \textbf{0.845} & \textbf{0.830} & 0.810 & 0.820 & 0.830 & 0.780 & 0.810 & 0.700 \\

\bottomrule
\end{tabular}
\vspace{-10pt}
\label{tab:tom_combined}
\end{table*}
%


\section{Conclusion}
We introduced a structured cognitive trajectory framework for Theory-of-Mind reasoning that models mental states as dynamic belief graphs with multiple interacting beliefs jointly evolving over time. Our framework  integrates LLM-derived semantic representations with structured probabilistic latent variable modeling to learn temporally coherent multi-belief dynamics and belief-conditioned decision-making from observation-action trajectories without belief supervision. Across real-world wildfire evacuation datasets containing multi-step multi-belief and standard ToM benchmarks, our approach improves action prediction while recovering interpretable latent belief trajectories and human-like belief interaction structures without using any belief ground truth. 
These results show that structured latent belief modeling provides a promising direction for accurate and behaviorally grounded ToM reasoning beyond static or independent belief representations.

\noindent\textbf{Limitations and Future Work:} Existing ToM benchmarks provide only short and weakly temporally grounded trajectories, limiting our extensive evaluation of long-horizon belief evolution and reasoning consistency.  Future work includes extending the framework to richer multimodal settings, improving scalability for larger belief spaces, and exploring enhancing LLM alignment through structured latent belief reasoning.


\bibliographystyle{plainnat}
\bibliography{example_paper}

@misc{zhao2021calibrate,
      title={Calibrate Before Use: Improving Few-Shot Performance of Language Models}, 
      author={Tony Z. Zhao and Eric Wallace and Shi Feng and Dan Klein and Sameer Singh},
      year={2021},
      eprint={2102.09690},
      archivePrefix={arXiv},
      primaryClass={cs.CL},
      url={https://arxiv.org/abs/2102.09690}, 
}

@inproceedings{
valmeekam2023on,
title={On the Planning Abilities of Large Language Models - A Critical Investigation},
author={Karthik Valmeekam and Matthew Marquez and Sarath Sreedharan and Subbarao Kambhampati},
booktitle={Thirty-seventh Conference on Neural Information Processing Systems},
year={2023},
url={https://openreview.net/forum?id=X6dEqXIsEW}
}

@misc{shi2023l,
      title={Large Language Models Can Be Easily Distracted by Irrelevant Context}, 
      author={Freda Shi and Xinyun Chen and Kanishka Misra and Nathan Scales and David Dohan and Ed Chi and Nathanael Schärli and Denny Zhou},
      year={2023},
      eprint={2302.00093},
      archivePrefix={arXiv},
      primaryClass={cs.CL},
      url={https://arxiv.org/abs/2302.00093}, 
}

@article{liu-etal-2024,
    title = "Lost in the Middle: How Language Models Use Long Contexts",
    author = "Liu, Nelson F.  and
      Lin, Kevin  and
      Hewitt, John  and
      Paranjape, Ashwin  and
      Bevilacqua, Michele  and
      Petroni, Fabio  and
      Liang, Percy",
    journal = "Transactions of the Association for Computational Linguistics",
    volume = "12",
    year = "2024",
    address = "Cambridge, MA",
    publisher = "MIT Press",
    doi = "10.1162/tacl_a_00638",
    pages = "157--173",
}

@misc{ouyang2022,
      title={Training language models to follow instructions with human feedback}, 
      author={Long Ouyang and Jeff Wu and Xu Jiang and Diogo Almeida and Carroll L. Wainwright and Pamela Mishkin and Chong Zhang and Sandhini Agarwal and Katarina Slama and Alex Ray and John Schulman and Jacob Hilton and Fraser Kelton and Luke Miller and Maddie Simens and Amanda Askell and Peter Welinder and Paul Christiano and Jan Leike and Ryan Lowe},
      year={2022},
      eprint={2203.02155},
      archivePrefix={arXiv},
      primaryClass={cs.CL},
      url={https://arxiv.org/abs/2203.02155}, 
}

@misc{kingma2022auto,
      title={Auto-Encoding Variational Bayes}, 
      author={Diederik P Kingma and Max Welling},
      year={2022},
      eprint={1312.6114},
      archivePrefix={arXiv},
      primaryClass={stat.ML},
      url={https://arxiv.org/abs/1312.6114}, 
}

@article{Lyu2022,
  author    = {Lyu, Yiping and Adams, Thomas},
  title     = {Preparing for real-time weather risk management: the decision models of household evacuation under uncertainty for Taiwanese and US residents},
  journal   = {Natural Hazards},
  volume    = {114},
  pages     = {405--425},
  year      = {2022},
  doi       = {10.1007/s11069-022-05395-8}
}

@article{Kinateder2015,
  author    = {Kinateder, Markus T. and Kuligowski, Erica D. and Reneke, Paul A. and others},
  title     = {Risk perception in fire evacuation behavior revisited: definitions, related concepts, and empirical evidence},
  journal   = {Fire Safety Science Review},
  volume    = {4},
  number    = {1},
  year      = {2015},
  doi       = {10.1186/s40038-014-0005-z}
}

@article{Kinsey2019,
  author    = {Kinsey, Michael J. and Gwynne, Steven M. V. and Kuligowski, Erica D. and others},
  title     = {Cognitive Biases Within Decision Making During Fire Evacuations},
  journal   = {Fire Technology},
  volume    = {55},
  pages     = {465--485},
  year      = {2019},
  doi       = {10.1007/s10694-018-0708-0}
}

@inbook{Cova2024,
author = {Cova, Thomas and Drews, Frank},
year = {2024},
month = {05},
pages = {237-250},
publisher = {Routledge},
title = {Wildfire Protective Actions and Collective Spatial Cognition},
doi = {10.4324/9781003202738-11}
}

@article{Alfiana2025,
author = {Alfiana, Alfiana and Azizi, Muhammad and Primafira, Andi and Prayana, I and Ahmad, Srifatmawati},
year = {2025},
month = {03},
pages = {941-946},
title = {BEHAVIORAL FINANCE AND ITS IMPACT ON CORPORATE FINANCIAL DECISION MAKING},
volume = {8},
journal = {Journal of Economic, Bussines and Accounting (COSTING)},
doi = {10.31539/costing.v8i2.14394}
}

@article{Croskerry2009,
  title={A universal model of diagnostic reasoning},
  author={Croskerry, Pat},
  journal={Academic Medicine},
  volume={84},
  number={8},
  pages={1022--1028},
  year={2009},
  publisher={LWW}
}

@article{Lakens2013,
  author    = {Lakens, Dani{\"e}l},
  title     = {Calculating and reporting effect sizes to facilitate cumulative science: a practical primer for t-tests and ANOVAs},
  journal   = {Frontiers in Psychology},
  volume    = {4},
  pages     = {863},
  year      = {2013},
  month     = nov,
  doi       = {10.3389/fpsyg.2013.00863},
  pmid      = {24324449},
  pmcid     = {PMC3840331},
}

@misc{kingma2022,
      title={Auto-Encoding Variational Bayes}, 
      author={Diederik P Kingma and Max Welling},
      year={2022},
      eprint={1312.6114},
      archivePrefix={arXiv},
      primaryClass={stat.ML},
      url={https://arxiv.org/abs/1312.6114}, 
}

@inproceedings{Rezende2014,
author = {Rezende, Danilo Jimenez and Mohamed, Shakir and Wierstra, Daan},
title = {Stochastic backpropagation and approximate inference in deep generative models},
year = {2014},
publisher = {JMLR.org},
booktitle = {Proceedings of the 31st International Conference on International Conference on Machine Learning - Volume 32},
pages = {II–1278–II–1286},
location = {Beijing, China},
series = {ICML'14}
}

@misc{vaswani2023,
      title={Attention Is All You Need}, 
      author={Ashish Vaswani and Noam Shazeer and Niki Parmar and Jakob Uszkoreit and Llion Jones and Aidan N. Gomez and Lukasz Kaiser and Illia Polosukhin},
      year={2023},
      eprint={1706.03762},
      archivePrefix={arXiv},
      primaryClass={cs.CL},
      url={https://arxiv.org/abs/1706.03762}, 
}

@inproceedings{LeCun2006ATO,
  title={A Tutorial on Energy-Based Learning},
  author={Yann LeCun and Sumit Chopra and Raia Hadsell and Aurelio Ranzato and Fu Jie Huang},
  year={2006},
  url={https://api.semanticscholar.org/CorpusID:8531544}
}

@article{PASTEUNING2025119833,
title = {Dynamic time warping to model daily life stress reactivity in a clinical and non-clinical sample – An ecological momentary assessment study},
journal = {Journal of Affective Disorders},
volume = {390},
pages = {119833},
year = {2025},
issn = {0165-0327},
doi = {https://doi.org/10.1016/j.jad.2025.119833},
author = {Jasmin M. Pasteuning and Caroline Broeder and Milou S.C. Sep and Bernet M. Elzinga and Brenda W.J.H. Penninx and Christiaan H. Vinkers and Erik J. Giltay},
}

@article{David2019,
author = {David C. Funder and Daniel J. Ozer},
title ={Evaluating Effect Size in Psychological Research: Sense and Nonsense},

journal = {Advances in Methods and Practices in Psychological Science},
volume = {2},
number = {2},
pages = {156-168},
year = {2019},
doi = {10.1177/2515245919847202}
}

@ARTICLE{Zhou2013,
  author={Zhou, Feng and De la Torre, Fernando and Hodgins, Jessica K.},
  journal={IEEE Transactions on Pattern Analysis and Machine Intelligence}, 
  title={Hierarchical Aligned Cluster Analysis for Temporal Clustering of Human Motion}, 
  year={2013},
  volume={35},
  number={3},
  pages={582-596},
  doi={10.1109/TPAMI.2012.137}}

@inproceedings{Keles2021,
  title={Efﬁcient prediction of trait judgments from faces using deep neural networks},
  author={Umit Keles and C. L. Lin and Ralph Adolphs},
  year={2021},
  url={https://api.semanticscholar.org/CorpusID:236746158}
}

@article{steyvers2025llmknowledge,
  author    = {Mark Steyvers and Hector Tejeda and Aviral Kumar and others},
  title     = {What large language models know and what people think they know},
  journal   = {Nature Machine Intelligence},
  volume    = {7},
  pages     = {221--231},
  year      = {2025},
  doi       = {10.1038/s42256-024-00976-7},
  url       = {https://doi.org/10.1038/s42256-024-00976-7}
}

@inproceedings{sclar-etal-2023-minding,
    title = "Minding Language Models' (Lack of) Theory of Mind: A Plug-and-Play Multi-Character Belief Tracker",
    author = "Sclar, Melanie  and
      Kumar, Sachin  and
      West, Peter  and
      Suhr, Alane  and
      Choi, Yejin  and
      Tsvetkov, Yulia",
    editor = "Rogers, Anna  and
      Boyd-Graber, Jordan  and
      Okazaki, Naoaki",
    booktitle = "Proceedings of the 61st Annual Meeting of the Association for Computational Linguistics (Volume 1: Long Papers)",
    month = jul,
    year = "2023",
    address = "Toronto, Canada",
    publisher = "Association for Computational Linguistics",
    url = "https://aclanthology.org/2023.acl-long.780/",
    doi = "10.18653/v1/2023.acl-long.780",
    pages = "13960--13980"
}

@inproceedings{wilf-etal-2024-think,
    title = "Think Twice: Perspective-Taking Improves Large Language Models' Theory-of-Mind Capabilities",
    author = "Wilf, Alex  and
      Lee, Sihyun  and
      Liang, Paul Pu  and
      Morency, Louis-Philippe",
    editor = "Ku, Lun-Wei  and
      Martins, Andre  and
      Srikumar, Vivek",
    booktitle = "Proceedings of the 62nd Annual Meeting of the Association for Computational Linguistics (Volume 1: Long Papers)",
    month = aug,
    year = "2024",
    address = "Bangkok, Thailand",
    publisher = "Association for Computational Linguistics",
    url = "https://aclanthology.org/2024.acl-long.451/",
    doi = "10.18653/v1/2024.acl-long.451",
    pages = "8292--8308"
}

@article{JARAETTINGER2019105,
title = {Theory of mind as inverse reinforcement learning},
journal = {Current Opinion in Behavioral Sciences},
volume = {29},
pages = {105-110},
year = {2019},
note = {Artificial Intelligence},
issn = {2352-1546},
doi = {https://doi.org/10.1016/j.cobeha.2019.04.010},
url = {https://www.sciencedirect.com/science/article/pii/S2352154618302055},
author = {Julian Jara-Ettinger}
}

@article{Christian2024,
  author    = {Christian, Philipp and Kaiser, Jan and Taylor, Peter C. and George, M. and Sch{\"u}tz-Bosbach, Simone and Soutschek, Alexander},
  title     = {Belief Updating during Social Interactions: Neural Dynamics and Causal Role of Dorsomedial Prefrontal Cortex},
  journal   = {The Journal of Neuroscience},
  year      = {2024},
  volume    = {44},
  number    = {22},
  pages     = {e1669232024},
  doi       = {10.1523/JNEUROSCI.1669-23.2024},
  pmid      = {38649270},
  pmcid     = {PMC11140663}
}

@article{Nan2024,
author={Nan,Jason and Herbert,Matthew S. and Purpura,Suzanna and Henneken,Andrea N. and Ramanathan,Dhakshin and Mishra,Jyoti},
year={2024},
title={Personalized Machine Learning-Based Prediction of Wellbeing and Empathy in Healthcare Professionals},
journal={Sensors},
volume={24},
number={8},
pages={2640},
language={English},
}

@inproceedings{
tang2025model,
title={Model Reconciliation via Cost-Optimal Explanations in Probabilistic Logic Programming},
author={Yinxu Tang and Stylianos Loukas Vasileiou and Vincent Derkinderen and William Yeoh},
booktitle={The Thirty-ninth Annual Conference on Neural Information Processing Systems},
year={2025},
}

@misc{zhang2025personalizat,
      title={Personalization of Large Language Models: A Survey}, 
      author={Zhehao Zhang and Ryan A. Rossi and Branislav Kveton and Yijia Shao and Diyi Yang and Hamed Zamani and Franck Dernoncourt and Joe Barrow and Tong Yu and Sungchul Kim and Ruiyi Zhang and Jiuxiang Gu and Tyler Derr and Hongjie Chen and Junda Wu and Xiang Chen and Zichao Wang and Subrata Mitra and Nedim Lipka and Nesreen Ahmed and Yu Wang},
      year={2025},
      eprint={2411.00027},
      archivePrefix={arXiv},
      primaryClass={cs.CL},
      url={https://arxiv.org/abs/2411.00027}, 
}

@inproceedings{chen-etal-2025,
    title = "From Perceptions to Decisions: Wildfire Evacuation Decision Prediction with Behavioral Theory-informed {LLM}s",
    author = "Chen, Ruxiao  and
      Wang, Chenguang  and
      Sun, Yuran  and
      Zhao, Xilei  and
      Xu, Susu",
    editor = "Che, Wanxiang  and
      Nabende, Joyce  and
      Shutova, Ekaterina  and
      Pilehvar, Mohammad Taher",
    booktitle = "Proceedings of the 63rd Annual Meeting of the Association for Computational Linguistics (Volume 1: Long Papers)",
    month = jul,
    year = "2025",
    address = "Vienna, Austria",
    publisher = "Association for Computational Linguistics",
    doi = "10.18653/v1/2025.acl-long.1438",
    pages = "29754--29778",
    ISBN = "979-8-89176-251-0",
}

@article{KULIGOWSKI2022105541,
title = {Modeling evacuation decisions in the 2019 Kincade fire in California},
journal = {Safety Science},
volume = {146},
pages = {105541},
year = {2022},
issn = {0925-7535},
doi = {https://doi.org/10.1016/j.ssci.2021.105541},
url = {https://www.sciencedirect.com/science/article/pii/S0925753521003842},
author = {Erica D. Kuligowski and Xilei Zhao and Ruggiero Lovreglio and Ningzhe Xu and Kaitai Yang and Aaron Westbury and Daniel Nilsson and Nancy Brown}
}

@article{FORRISTER2024100729,
title = {Analyzing Risk Perception, Evacuation Decision and Delay Time: A Case Study of the 2021 Marshall Fire in Colorado},
journal = {Travel Behaviour and Society},
volume = {35},
pages = {100729},
year = {2024},
issn = {2214-367X},
doi = {https://doi.org/10.1016/j.tbs.2023.100729},
url = {https://www.sciencedirect.com/science/article/pii/S2214367X23001801},
author = {Ana Forrister and Erica D. Kuligowski and Yuran Sun and Xiang Yan and Ruggiero Lovreglio and Thomas J. Cova and Xilei Zhao}
}

@misc{belanger2016,
      title={Structured Prediction Energy Networks}, 
      author={David Belanger and Andrew McCallum},
      year={2016},
      eprint={1511.06350},
      archivePrefix={arXiv},
      primaryClass={cs.LG},
      url={https://arxiv.org/abs/1511.06350}, 
}

@misc{krishnan2016,
      title={Structured Inference Networks for Nonlinear State Space Models}, 
      author={Rahul G. Krishnan and Uri Shalit and David Sontag},
      year={2016},
      eprint={1609.09869},
      archivePrefix={arXiv},
      primaryClass={stat.ML},
      url={https://arxiv.org/abs/1609.09869}, 
}

@misc{zhang2023,
      title={Factor Graph Neural Networks}, 
      author={Zhen Zhang and Mohammed Haroon Dupty and Fan Wu and Javen Qinfeng Shi and Wee Sun Lee},
      year={2023},
      eprint={2308.00887},
      archivePrefix={arXiv},
      primaryClass={cs.LG},
      url={https://arxiv.org/abs/2308.00887}, 
}

@inproceedings{chung2015,
 author = {Chung, Junyoung and Kastner, Kyle and Dinh, Laurent and Goel, Kratarth and Courville, Aaron C and Bengio, Yoshua},
 booktitle = {Advances in Neural Information Processing Systems},
 editor = {C. Cortes and N. Lawrence and D. Lee and M. Sugiyama and R. Garnett},
 pages = {},
 publisher = {Curran Associates, Inc.},
 title = {A Recurrent Latent Variable Model for Sequential Data},
 url = {https://proceedings.neurips.cc/paper_files/paper/2015/file/b618c3210e934362ac261db280128c22-Paper.pdf},
 volume = {28},
 year = {2015}
}

@inproceedings{
zhang2025autotom,
title={AutoToM: Scaling Model-based Mental Inference via Automated Agent Modeling},
author={Zhining Zhang and Chuanyang Jin and Mung Yao Jia and Shunchi Zhang and Tianmin Shu},
booktitle={The Thirty-ninth Annual Conference on Neural Information Processing Systems},
year={2025},
}

@article{PADM2012,
author = {Lindell, Michael K. and Perry, Ronald W.},
title = {The Protective Action Decision Model: Theoretical Modifications and Additional Evidence},
journal = {Risk Analysis},
volume = {32},
number = {4},
pages = {616-632},
doi = {https://doi.org/10.1111/j.1539-6924.2011.01647.x},

year = {2012}
}

@article{champ2016,
author = {Champ, Patricia A and Brenkert-Smith, Hannah},
title = {Is Seeing Believing? Perceptions of Wildfire Risk Over Time},
journal = {Risk Analysis},
volume = {36},
number = {4},
pages = {816-830},
doi = {https://doi.org/10.1111/risa.12465},

year = {2016}
}

@misc{jin2024mmtom,
      title={MMToM-QA: Multimodal Theory of Mind Question Answering}, 
      author={Chuanyang Jin and Yutong Wu and Jing Cao and Jiannan Xiang and Yen-Ling Kuo and Zhiting Hu and Tomer Ullman and Antonio Torralba and Joshua B. Tenenbaum and Tianmin Shu},
      year={2024},
      eprint={2401.08743},
      archivePrefix={arXiv},
      primaryClass={cs.AI},
}

@misc{shi2025mumatom,
      title={MuMA-ToM: Multi-modal Multi-Agent Theory of Mind}, 
      author={Haojun Shi and Suyu Ye and Xinyu Fang and Chuanyang Jin and Leyla Isik and Yen-Ling Kuo and Tianmin Shu},
      year={2025},
      eprint={2408.12574},
      archivePrefix={arXiv},
      primaryClass={cs.AI},
}

@article{Jha2024, title={Neural Amortized Inference for Nested Multi-Agent Reasoning}, volume={38}, 
DOI={10.1609/aaai.v38i1.27808}, 
number={1}, 
journal={Proceedings of the AAAI Conference on Artificial Intelligence}, author={Jha, Kunal and Le, Tuan Anh and Jin, Chuanyang and Kuo, Yen-Ling and Tenenbaum, Joshua B. and Shu, Tianmin}, 
year={2024}, 
month={Mar.}, 
pages={530-537} }

@article{Baker2017Rational,
  title   = {Rational quantitative attribution of beliefs, desires and percepts in human mentalizing},
  author  = {Baker, Chris L. and Jara-Ettinger, Julian and Saxe, Rebecca and Tenenbaum, Joshua B.},
  journal = {Nature Human Behaviour},
  volume  = {1},
  pages   = {0064},
  year    = {2017},
  doi     = {10.1038/s41562-017-0064}
}

@inproceedings{
kim2025hypo,
title={Hypothesis-Driven Theory-of-Mind Reasoning for Large Language Models},
author={Hyunwoo Kim and Melanie Sclar and Tan Zhi-Xuan and Lance Ying and Sydney Levine and Yang Liu and Joshua B. Tenenbaum and Yejin Choi},
booktitle={Second Conference on Language Modeling},
year={2025},
url={https://openreview.net/forum?id=yGQqTuSJPK}
}

@misc{yang2025,
      title={Large Language Models as Theory of Mind Aware Generative Agents with Counterfactual Reflection}, 
      author={Bo Yang and Jiaxian Guo and Yusuke Iwasawa and Yutaka Matsuo},
      year={2025},
      eprint={2501.15355},
      archivePrefix={arXiv},
      primaryClass={cs.CL},
      url={https://arxiv.org/abs/2501.15355}, 
}

@article{Paton2019DisasterRisk,
  author    = {Paton, Douglas},
  title     = {Disaster risk reduction: Psychological perspectives on preparedness},
  journal   = {Australian Journal of Psychology},
  volume    = {71},
  number    = {4},
  pages     = {327--341},
  year      = {2019},
  doi       = {10.1111/ajpy.12237}
}

@article{WIMMER1983103,
title = {Beliefs about beliefs: Representation and constraining function of wrong beliefs in young children's understanding of deception},
journal = {Cognition},
volume = {13},
number = {1},
pages = {103-128},
year = {1983},
author = {Heinz Wimmer and Josef Perner},
}

@article{tom2009,
author = {Obiols, Jordi and Berrios, German},
year = {2009},
month = {09},
pages = {377-92},
title = {The historical roots of Theory of Mind: The work of James Mark Baldwin},
volume = {20},
journal = {History of psychiatry},
doi = {10.1177/0957154X08337334}
}

@misc{hou2025egoso,
      title={EgoSocialArena: Benchmarking the Social Intelligence of Large Language Models from a First-person Perspective}, 
      author={Guiyang Hou and Wenqi Zhang and Yongliang Shen and Zeqi Tan and Sihao Shen and Weiming Lu},
      year={2025},
      eprint={2410.06195},
      archivePrefix={arXiv},
      primaryClass={cs.CL},
      url={https://arxiv.org/abs/2410.06195}, 
}

@article{kinsey2019cognitive,
  title   = {Cognitive Biases Within Decision Making During Fire Evacuations},
  author  = {Kinsey, Michael J. and Gwynne, Steven M. V. and Kuligowski, Erica D. and others},
  journal = {Fire Technology},
  volume  = {55},
  number  = {2},
  pages   = {465--485},
  year    = {2019},
  doi     = {10.1007/s10694-018-0708-0},
  publisher = {Springer}
}

@inproceedings{kim-etal-2023-fantom,
    title = "{FANT}o{M}: A Benchmark for Stress-testing Machine Theory of Mind in Interactions",
    author = "Kim, Hyunwoo  and
      Sclar, Melanie  and
      Zhou, Xuhui  and
      Bras, Ronan  and
      Kim, Gunhee  and
      Choi, Yejin  and
      Sap, Maarten",
    editor = "Bouamor, Houda  and
      Pino, Juan  and
      Bali, Kalika",
    booktitle = "Proceedings of the 2023 Conference on Empirical Methods in Natural Language Processing",
    month = dec,
    year = "2023",
    address = "Singapore",
    publisher = "Association for Computational Linguistics",
    url = "https://aclanthology.org/2023.emnlp-main.890/",
    doi = "10.18653/v1/2023.emnlp-main.890",
    pages = "14397--14413"
}

@inproceedings{le-etal-2019-revisiting,
    title = "Revisiting the Evaluation of Theory of Mind through Question Answering",
    author = "Le, Matthew  and
      Boureau, Y-Lan  and
      Nickel, Maximilian",
    editor = "Inui, Kentaro  and
      Jiang, Jing  and
      Ng, Vincent  and
      Wan, Xiaojun",
    booktitle = "Proceedings of the 2019 Conference on Empirical Methods in Natural Language Processing and the 9th International Joint Conference on Natural Language Processing (EMNLP-IJCNLP)",
    month = nov,
    year = "2019",
    address = "Hong Kong, China",
    publisher = "Association for Computational Linguistics",
    url = "https://aclanthology.org/D19-1598/",
    doi = "10.18653/v1/D19-1598",
    pages = "5872--5877"
}

@inproceedings{NEURIPS2023_2b9efb08,
 author = {Gandhi, Kanishk and Fraenken, Jan-Philipp and Gerstenberg, Tobias and Goodman, Noah},
 booktitle = {Advances in Neural Information Processing Systems},
 editor = {A. Oh and T. Naumann and A. Globerson and K. Saenko and M. Hardt and S. Levine},
 pages = {13518--13529},
 publisher = {Curran Associates, Inc.},
 title = {Understanding Social Reasoning in Language Models with Language Models},
 url = {https://proceedings.neurips.cc/paper_files/paper/2023/file/2b9efb085d3829a2aadffab63ba206de-Paper-Datasets_and_Benchmarks.pdf},
 volume = {36},
 year = {2023}
}

@misc{xu2024opentomcomprehen,
      title={OpenToM: A Comprehensive Benchmark for Evaluating Theory-of-Mind Reasoning Capabilities of Large Language Models}, 
      author={Hainiu Xu and Runcong Zhao and Lixing Zhu and Jinhua Du and Yulan He},
      year={2024},
      eprint={2402.06044},
      archivePrefix={arXiv},
      primaryClass={cs.AI},
      url={https://arxiv.org/abs/2402.06044}, 
}

@inproceedings{Schulhoff2024ThePR,
  title={The Prompt Report: A Systematic Survey of Prompt Engineering Techniques},
  author={Sander Schulhoff and Michael Ilie and Nishant Balepur and Konstantine Kahadze and Amanda Liu and Chenglei Si and Yinheng Li and Aayush Gupta and HyoJung Han and Sevien Schulhoff and Pranav Sandeep Dulepet and Saurav Vidyadhara and Dayeon Ki and Sweta Agrawal and Chau Pham and Gerson C. Kroiz and Feileen Li and Hudson Tao and Ashay Srivastava and Hevander Da Costa and Saloni Gupta and Megan L. Rogers and Inna Goncearenco and Giuseppe Sarli and Igor Galynker and Denis Peskoff and Marine Carpuat and Jules White and Shyamal Anadkat and Alexander Hoyle and Philip Resnik},
  year={2024},
  url={https://api.semanticscholar.org/CorpusID:276617957}
}

@inproceedings{xia-etal-2025-influences,
    title = "Influences on {LLM} Calibration: A Study of Response Agreement, Loss Functions, and Prompt Styles",
    author = "Xia, Yuxi  and
      Luz De Araujo, Pedro Henrique  and
      Zaporojets, Klim  and
      Roth, Benjamin",
    editor = "Che, Wanxiang  and
      Nabende, Joyce  and
      Shutova, Ekaterina  and
      Pilehvar, Mohammad Taher",
    booktitle = "Proceedings of the 63rd Annual Meeting of the Association for Computational Linguistics (Volume 1: Long Papers)",
    month = jul,
    year = "2025",
    address = "Vienna, Austria",
    publisher = "Association for Computational Linguistics",
    url = "https://aclanthology.org/2025.acl-long.188/",
    doi = "10.18653/v1/2025.acl-long.188",
    pages = "3740--3761",
    ISBN = "979-8-89176-251-0"
}

@inproceedings{huang-etal-2025-path,
    title = "Path Drift in Large Reasoning Models: How First-Person Commitments Override Safety",
    author = "Huang, Yuyi  and
      Zhan, Runzhe  and
      Chao, Lidia S.  and
      Tao, Ailin  and
      Wong, Derek F.",
    editor = "Christodoulopoulos, Christos  and
      Chakraborty, Tanmoy  and
      Rose, Carolyn  and
      Peng, Violet",
    booktitle = "Proceedings of the 2025 Conference on Empirical Methods in Natural Language Processing",
    month = nov,
    year = "2025",
    address = "Suzhou, China",
    publisher = "Association for Computational Linguistics",
    url = "https://aclanthology.org/2025.emnlp-main.990/",
    doi = "10.18653/v1/2025.emnlp-main.990",
    pages = "19602--19616",
    ISBN = "979-8-89176-332-6"
}

\clearpage

\appendix

\section{Metrics}
\label{appx:metrics}

This appendix details the quantitative metrics used to evaluate belief prediction quality, belief interaction structure, and temporal belief dynamics. All metrics are computed \emph{post hoc} for evaluation only and are not used for training.

\subsection{Standardized Mean Difference (Action-Conditioned Belief Change)}

Let $\mathbf{b}_{t} \in [0,1]^K$ denote the vector of belief marginals at time $t$, and let $a_t$ denote the discrete action at time $t$. We define the belief change magnitude between consecutive steps as
\begin{equation}
\Delta \mathbf{b}_t = \mathbf{b}_t - \mathbf{b}_{t-1},
\qquad
x_t = \|\Delta \mathbf{b}_t\|_1 = \sum_{i=1}^K \lvert b_{t,i} - b_{t-1,i} \rvert .
\end{equation}

We introduce an action-change indicator $y_t = \mathbb{I}[a_t \neq a_{t-1}]$, where $y_t=1$ indicates an action change and $y_t=0$ otherwise. To quantify whether belief updates are selectively amplified at action-changing timesteps, we compute the standardized mean difference (Cohen’s $d$):
\begin{equation}
d = \frac{\mu_{1} - \mu_{0}}{s_{\mathrm{pooled}}},
\end{equation}
where $\mu_{1} = \mathbb{E}[x_t \mid y_t=1]$ and $\mu_{0} = \mathbb{E}[x_t \mid y_t=0]$ are the conditional means of belief change magnitude, and
\begin{equation}
s_{\mathrm{pooled}} =
\sqrt{
\frac{(n_{1}-1)s_{1}^{2} + (n_{0}-1)s_{0}^{2}}
{n_{1}+n_{0}-2}
}
\end{equation}
is the pooled standard deviation computed from the within-group variances $s_1^2, s_0^2$ and sample sizes $n_1, n_0$.

A larger positive $d$ indicates that belief updates are stronger at timesteps where actions change, reflecting action-aligned belief dynamics rather than uniform temporal drift.

\subsection{Dynamic Time Warping (DTW)}

To measure trajectory-level alignment between predicted and ground-truth belief dynamics, we use Dynamic Time Warping (DTW).
Let $b^{\mathrm{pred}}_{t,i}$ and $b^{\mathrm{gt}}_{t,i}$ denote the predicted and ground-truth belief values for belief $i$ at time $t$, respectively. For each belief dimension $i$ and individual sequence, we compute the DTW distance
\begin{equation}
\mathrm{DTW}_i
=
\mathrm{DTW}\!\left(
\{ b^{\mathrm{pred}}_{t,i} \}_{t=1}^{T},
\{ b^{\mathrm{gt}}_{t,i} \}_{t=1}^{T}
\right),
\end{equation}
where $\mathrm{DTW}(\cdot,\cdot)$ denotes the standard dynamic time warping distance.

To account for sequence length, we normalize by the valid trajectory length and report the average DTW across beliefs and individuals:
\begin{equation}
\mathrm{DTW}_{\mathrm{avg}}
=
\mathbb{E}_{i,n}\!\left[ \frac{\mathrm{DTW}_{i,n}}{T_{n}} \right],
\end{equation}
where $n$ indexes individuals and $T_n$ is the valid length of sequence $n$.
Lower DTW indicates better temporal alignment between predicted and ground-truth belief trajectories.

\subsection{Spearman Rank Correlation}
\label{appx:spearman}

Given two variables $\{x_i\}_{i=1}^N$ and $\{y_i\}_{i=1}^N$, let
$\{R_i\}$ and $\{S_i\}$ denote their rank-transformed values.
The Spearman correlation coefficient is defined as
\begin{equation}
\rho
=
\frac{\sum_{i=1}^{N}(R_i-\bar{R})(S_i-\bar{S})}
{\sqrt{\sum_{i=1}^{N}(R_i-\bar{R})^2}\sqrt{\sum_{i=1}^{N}(S_i-\bar{S})^2}},
\end{equation}
where $\bar{R}$ and $\bar{S}$ are the mean ranks.
\clearpage
\section{Derivation of the Evidence Lower Bound (ELBO)}
\label{appx:elbo}

In this section, we provide the detailed derivation of the training objective ELBO \cite{kingma2022auto}. We formulate the objective by decomposing the sequence likelihood into time-step specific contributions. Specifically, we derive the ELBO for a single time step $t$, conditioned on the history (encapsulated in the previous belief state $\mathbf{b}_{t-1}$) and the current observation $o_t$.

\subsection{Problem Definition}
Let $\mathbf{b}_t$ denote the discrete latent belief vector, $a_t$ denote the observed action, and $o_t$ denote the observation at time $t$. The generative model factorizes the joint distribution at step $t$ as:
\begin{equation}
    p_\theta(a_t, \mathbf{b}_t \mid \mathbf{b}_{t-1}, o_t) = p_\theta(a_t \mid \mathbf{b}_t) \, p_\theta(\mathbf{b}_t \mid \mathbf{b}_{t-1}, o_t),
\end{equation}
where $p_\theta(\mathbf{b}_t \mid \mathbf{b}_{t-1}, o_t)$ is the belief transition prior and $p_\theta(a_t \mid \mathbf{b}_t)$ is the action likelihood.

We aim to maximize the conditional log-likelihood of the observed action $a_t$ given the previous state:
\begin{equation}
    \log p_\theta(a_t \mid \mathbf{b}_{t-1}, o_t) = \log \sum_{\mathbf{b}_t} p_\theta(a_t, \mathbf{b}_t \mid \mathbf{b}_{t-1}, o_t).
\end{equation}

\subsection{Derivation via Jensen's Inequality}
We introduce a variational approximation $q_\phi(\mathbf{b}_t \mid o_t, a_t)$ to approximate the posterior distribution of the beliefs. Rewriting the log-likelihood by multiplying and dividing by $q_\phi$:
\begin{equation}
    \log p_\theta(a_t \mid \mathbf{b}_{t-1}, o_t) = \log \sum_{\mathbf{b}_t} q_\phi(\mathbf{b}_t \mid o_t, a_t) \frac{p_\theta(a_t, \mathbf{b}_t \mid \mathbf{b}_{t-1}, o_t)}{q_\phi(\mathbf{b}_t \mid o_t, a_t)}.
\end{equation}
This summation is equivalent to the expectation under the variational distribution:
\begin{equation}
    \log p_\theta(a_t \mid \mathbf{b}_{t-1}, o_t) = \log \mathbb{E}_{\mathbf{b}_t \sim q_\phi} \left[ \frac{p_\theta(a_t, \mathbf{b}_t \mid \mathbf{b}_{t-1}, o_t)}{q_\phi(\mathbf{b}_t \mid o_t, a_t)} \right].
\end{equation}
Applying Jensen's Inequality ($\log \mathbb{E}[X] \ge \mathbb{E}[\log X]$) yields the Evidence Lower Bound (ELBO) for time step $t$:
\begin{equation}
    \log p_\theta(a_t \mid \mathbf{b}_{t-1}, o_t) \ge \mathbb{E}_{\mathbf{b}_t \sim q_\phi} \left[ \log \frac{p_\theta(a_t, \mathbf{b}_t \mid \mathbf{b}_{t-1}, o_t)}{q_\phi(\mathbf{b}_t \mid o_t, a_t)} \right].
\end{equation}

\subsection{Objective Function Decomposition}
We expand the joint probability in the numerator using the chain rule:
\begin{equation}
    \text{ELBO}_t = \mathbb{E}_{q_\phi} \left[ \log \frac{p_\theta(a_t \mid \mathbf{b}_t) \, p_\theta(\mathbf{b}_t \mid \mathbf{b}_{t-1}, o_t)}{q_\phi(\mathbf{b}_t \mid o_t, a_t)} \right].
\end{equation}
By separating the terms, we obtain the final objective function consisting of a reconstruction term and a regularization term:
\begin{equation}
\begin{aligned}
    \text{ELBO}_t &= \mathbb{E}_{q_\phi} \big[ \log p_\theta(a_t \mid \mathbf{b}_t) \big] + \mathbb{E}_{q_\phi} \left[ \log \frac{p_\theta(\mathbf{b}_t \mid \mathbf{b}_{t-1}, o_t)}{q_\phi(\mathbf{b}_t \mid o_t, a_t)} \right] \\
    &= \underbrace{\mathbb{E}_{q_\phi(\mathbf{b}_t \mid o_t, a_t)} \big[ \log p_\theta(a_t \mid \mathbf{b}_t) \big]}_{\mathcal{L}^{\text{act}}_t: \text{ Action Reconstruction}} 
    - \underbrace{D_{\text{KL}} \big( q_\phi(\mathbf{b}_t \mid o_t, a_t) \,\|\, p_\theta(\mathbf{b}_t \mid \mathbf{b}_{t-1}, o_t) \big)}_{\mathcal{L}^{\text{KL}}_t: \text{ Belief Consistency}}.
\end{aligned}
\end{equation}
The total training objective is the sum of these negative step-wise ELBOs over the sequence length $T$:
\begin{equation}
    \mathcal{L} = -\sum_{t=1}^{T} \text{ELBO}_t.
\end{equation}

\clearpage
\section{Results across different LLM Models}
\label{appx:allmodels}
We have extended our experiments to multiple LLM backbone including Qwen3-8B, Qwen3-32B, Qwen2.5-72B, LLaMA3-8B, LLaMA-3-70B, and DeepSeek-R1 70B.

\begin{figure}[H]
    \centering
    \includegraphics[width=\linewidth]{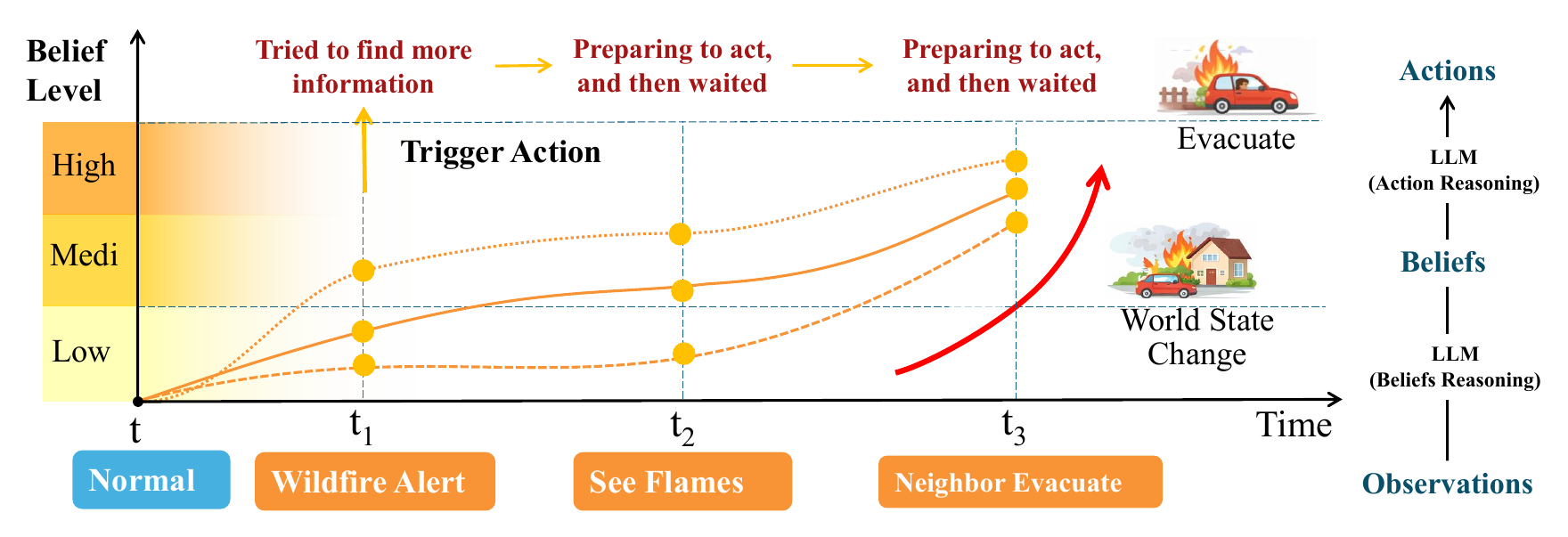}
    \caption{An example of how belief trajectories evolve with observations in high-stakes decision-making like wildfire evacuation, actions are triggerred upon threshold crossing.}
    \label{fig:overview}
\end{figure}

\begin{figure}[H]
    \centering
    \includegraphics[width=1\linewidth]{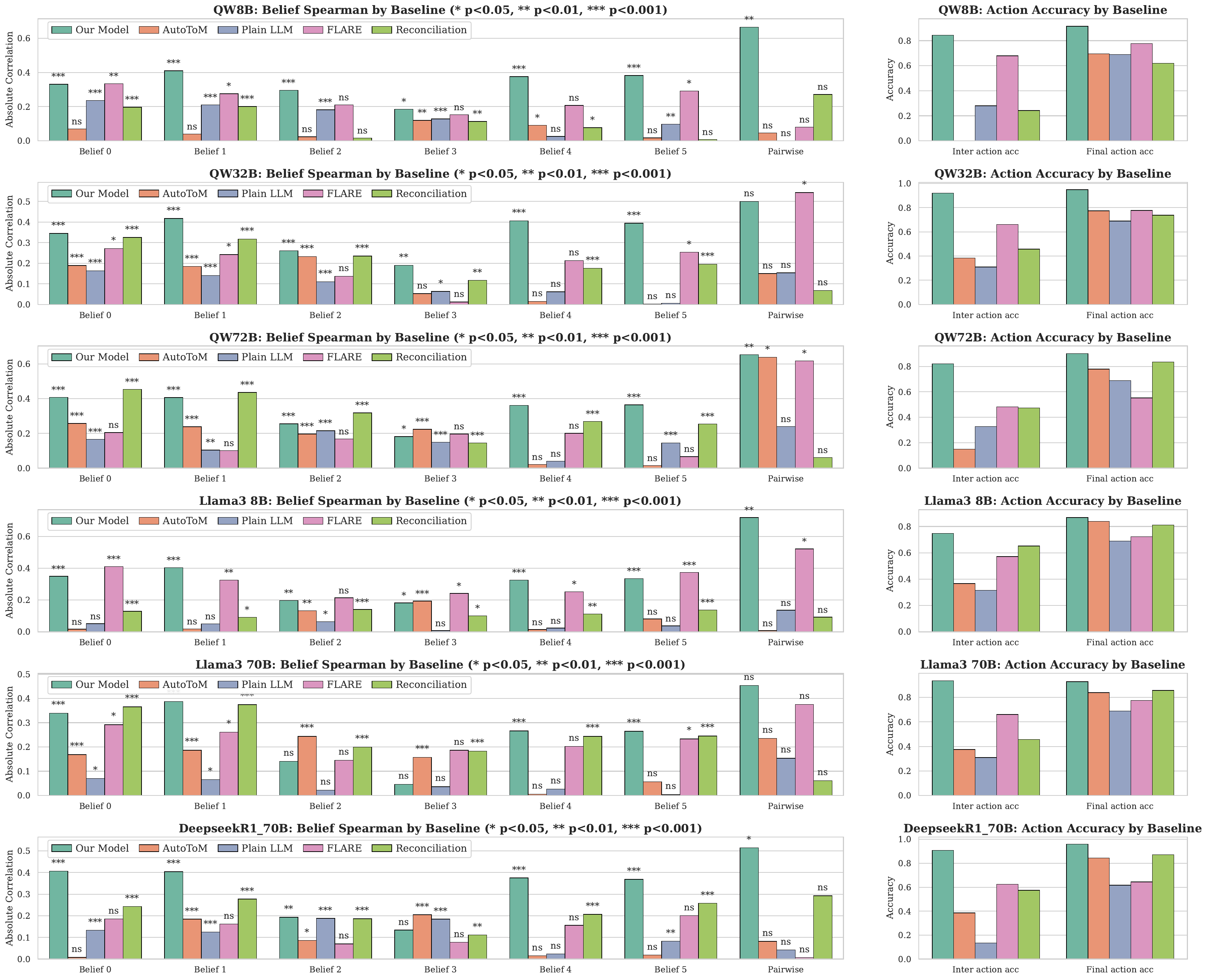}
    \caption{Results on wildfire dataset across all LLM backbone. Including belief spearman and action prediction accuracy}
    \label{fig:all_llm}
\end{figure}

\clearpage
\section{Prompts Sensitivity Analysis}
\label{appx:prompts_sen}
 Prior work has shown that prompt perspective alone can affect reasoning outcomes (first-person perspective and third-person perspective) \cite{huang-etal-2025-path}, and it is argued that first-person prompting better aligns with natural human cognition in social reasoning tasks \cite{hou2025egoso}. In disaster contexts, this is especially relevant due to optimism bias, where individuals underestimate their own risk while judging others as more vulnerable \cite{kinsey2019cognitive}. Motivated by this, our original setup adopts first-person prompting. 
Here, we further compare it with third-person prompting. In addition, prior work has shown that performance can vary across prompting strategies such as chain-of-thought (CoT) and few-shot prompting \cite{xia-etal-2025-influences,Schulhoff2024ThePR}. Therefore, we also evaluate CoT and few-shot prompting under the first-person setting to assess robustness to prompt design.

As shown in Table~\ref{tab:prompting}, compared to third-person prompting, first-person prompting consistently performs better in recovering belief structure, suggesting that adopting the agent’s perspective enables the LLM to more naturally reproduce human cognitive biases such as optimism bias. In contrast, third-person prompting weakens this alignment and leads to less stable belief estimates. Meanwhile, other prompting strategies (e.g., CoT, few-shot, probabilistic prompting) exhibit relatively minor and inconsistent differences, indicating that prompt format has limited impact compared to the choice of perspective.

\begin{table}[htbp]
\centering
\small
\setlength{\tabcolsep}{6pt} 
\renewcommand{\arraystretch}{1.3}
\begin{tabular}{>{\raggedright\arraybackslash}p{2.8cm}ccccc}
\toprule
 & \makecell{First\\person} 
 & \makecell{Third\\person} 
 & \makecell{Probabilistic\\prompting} 
 & \makecell{Few-shot\\prompting} 
 & \makecell{CoT\\prompting} \\
\midrule
Belief 0 & 0.3311 & 0.3207 & 0.3588 & 0.3638 & 0.3495 \\
Belief 1 & 0.4099 & 0.4241 & 0.4101 & 0.4021 & 0.4055 \\
Belief 2 & 0.2952 & 0.1046 & 0.2504 & 0.2478 & 0.2396 \\
Belief 3 & 0.1848 & 0.1029 & 0.2628 & 0.1747 & 0.2083 \\
Belief 4 & 0.3757 & 0.3329 & 0.3604 & 0.3614 & 0.3223 \\
Belief 5 & 0.3824 & 0.3241 & 0.3680 & 0.3551 & 0.3327 \\
\midrule
$p$-value 0 & $3.27\times10^{-6}$ & $6.83\times10^{-6}$ & $3.98\times10^{-7}$ & $2.67\times10^{-7}$ & $8.25\times10^{-7}$ \\
$p$-value 1 & $4.73\times10^{-9}$ & $1.20\times10^{-9}$ & $4.62\times10^{-9}$ & $9.76\times10^{-9}$ & $7.10\times10^{-9}$ \\
$p$-value 2 & $3.72\times10^{-5}$ & $1.52\times10^{-1}$ & $5.10\times10^{-4}$ & $5.85\times10^{-4}$ & $8.97\times10^{-4}$ \\
$p$-value 3 & $1.09\times10^{-2}$ & $1.59\times10^{-1}$ & $2.59\times10^{-4}$ & $1.62\times10^{-2}$ & $4.02\times10^{-3}$ \\
$p$-value 4 & $9.98\times10^{-8}$ & $2.86\times10^{-6}$ & $3.50\times10^{-7}$ & $3.24\times10^{-7}$ & $6.12\times10^{-6}$ \\
$p$-value 5 & $5.64\times10^{-8}$ & $5.37\times10^{-6}$ & $1.89\times10^{-7}$ & $5.34\times10^{-7}$ & $2.92\times10^{-6}$ \\
\midrule
Pairwise correlation & 0.6643 & 0.4071 & 0.7000 & 0.6179 & 0.4036 \\
Pairwise $p$-value & $6.91\times10^{-3}$ & $1.32\times10^{-1}$ & $3.67\times10^{-3}$ & $1.41\times10^{-2}$ & $1.36\times10^{-1}$ \\
\midrule
Intermediate action Acc & 0.8443 & 0.8439 & 0.8712 & 0.8932 & 0.8057 \\
Final action acc & 0.9157 & 0.8383 & 0.8567 & 0.9128 & 0.8735 \\
\bottomrule
\end{tabular}

\caption{Comparison across different prompting strategies. Compared on eliefs spearman correlations and action predictions accuracy.}
\label{tab:prompting}
\end{table}

\clearpage
\section{Beliefs Scalability Test}
\label{appx:scalability}

In our current setting, the number of belief dimensions K=6 is not arbitrarily chosen by us, but follows the behavioral theory, PADM, and PADM-guided survey design for the wildfire evacuation survey [9, 10]. To further analyze the effect of belief dimensionality, we conduct experiments with K=2,4,6,10,15.

For K<6, we use subsets of the original belief variables. As shown in Table ~\ref{tab:ablation_K}. When reducing the belief space to K=4, the model still preserves strong performance, while K=2 leads to a sharp drop, indicating that too few beliefs fail to capture sufficient decision-relevant information. Also, under K=2, belief estimation metrics appear higher. This is because the model is forced to rely on the limited belief variables to explain actions, and therefore must align them more closely with the true human beliefs.

When extending the belief space, a key challenge is that the number of joint belief configurations grows exponentially as $2^K$, which may lead to computational intractability for large $K$. To address this, we consider standard approximate inference methods, including mean-field and MCMC. As shown in Table~\ref{tab:inference_time}, exact inference remains efficient in our setting due to fully vectorized GPU computation, with only modest increases in runtime as $K$ grows. Meanwhile, mean-field provides the fastest approximation with small performance degradation, whereas MCMC is significantly slower due to its sequential nature and limited parallelizability on GPUs. These results indicate that while exact inference is sufficient for the current theory-grounded setting, the framework can be naturally extended to scalable approximate inference methods when $K$ becomes large.

\begin{table}[htbp]
\centering
\small
\setlength{\tabcolsep}{6pt}
\renewcommand{\arraystretch}{1.3}

\begin{tabular}{>{\raggedright\arraybackslash}p{3cm}ccc}
\toprule
 & 2 beliefs & 4 beliefs & Original Setup (6 beliefs) \\
\midrule
Belief 0 & - & 0.3187 ($7.83\times10^{-6}$) & 0.3311 ($3.27\times10^{-6}$) \\
Belief 1 & - & 0.4067 ($6.33\times10^{-9}$) & 0.4099 ($4.73\times10^{-9}$) \\
Belief 2 & 0.2637 ($2.47\times10^{-4}$) & 0.1795 ($1.35\times10^{-2}$) & 0.2952 ($3.72\times10^{-5}$) \\
Belief 3 & 0.3315 ($3.16\times10^{-6}$) & 0.1100 ($1.32\times10^{-1}$) & 0.1848 ($1.09\times10^{-2}$) \\
Belief 4 & - & - & 0.3757 ($9.98\times10^{-8}$) \\
Belief 5 & - & - & 0.3824 ($5.64\times10^{-8}$) \\
\midrule
Pairwise corr. & - & - & 0.6643 ($6.91\times10^{-3}$) \\
\midrule
Action acc & 0.6180 & 0.8349 & 0.8443 \\
Final action acc & 0.9041 & 0.9140 & 0.9157 \\
\bottomrule
\end{tabular}

\caption{Ablation study on the number of beliefs $K \in \{2,4,6\}$. Each entry reports Spearman correlation with $p$-value in parentheses.}
\label{tab:ablation_K}
\end{table}

\begin{table}[htbp]
\centering
\small
\setlength{\tabcolsep}{5pt}
\begin{tabular}{lcccc}
\toprule
Methods & Intermediate Acc & Final Acc & Inference Time (s) \\
\midrule
K=6 (Exact)      & 0.8443 & 0.9157 & 0.003031 \\
K=6 (MF)         & 0.8240 & 0.9167 & 0.001569 \\
K=6 (MCMC)       & 0.8223 & 0.9204 & 0.018991 \\
\midrule
K=10 (Exact)     & -      & -      & 0.003165 \\
K=10 (MF)        & -      & -      & 0.001895 \\
K=10 (MCMC)      & -      & -      & 0.031219 \\
\midrule
K=15 (Exact)     & -      & -      & 0.003718 \\
K=15 (MF)        & -      & -      & 0.002751 \\
K=15 (MCMC)      & -      & -      & 0.043742 \\
\bottomrule
\end{tabular}
\caption{Scaling the number of beliefs ($K=6,10,15$) and comparing approximation methods.}
\label{tab:inference_time}
\end{table}
\clearpage

\clearpage
\section{More Results}
\label{appx:more_results}
\begin{figure}[H]
    \centering
    \begin{subfigure}[t]{0.48\linewidth}
        \centering
        \includegraphics[width=\linewidth]{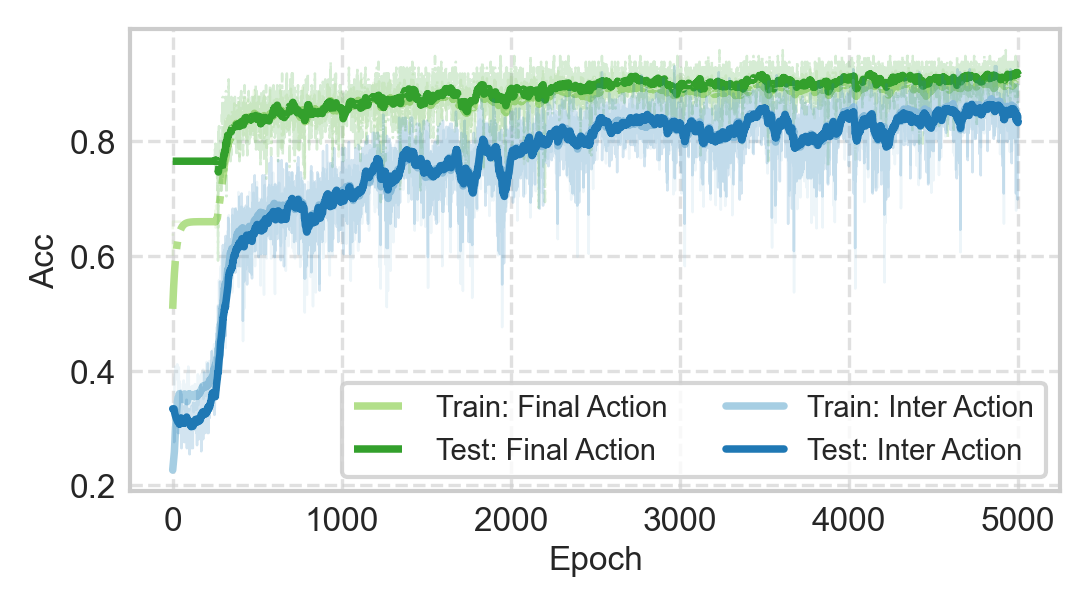}
        \caption{Action prediction accuracy.}
        \label{fig:action_acc}
    \end{subfigure}
    \hfill
    \begin{subfigure}[t]{0.48\linewidth}
        \centering
        \includegraphics[width=\linewidth]{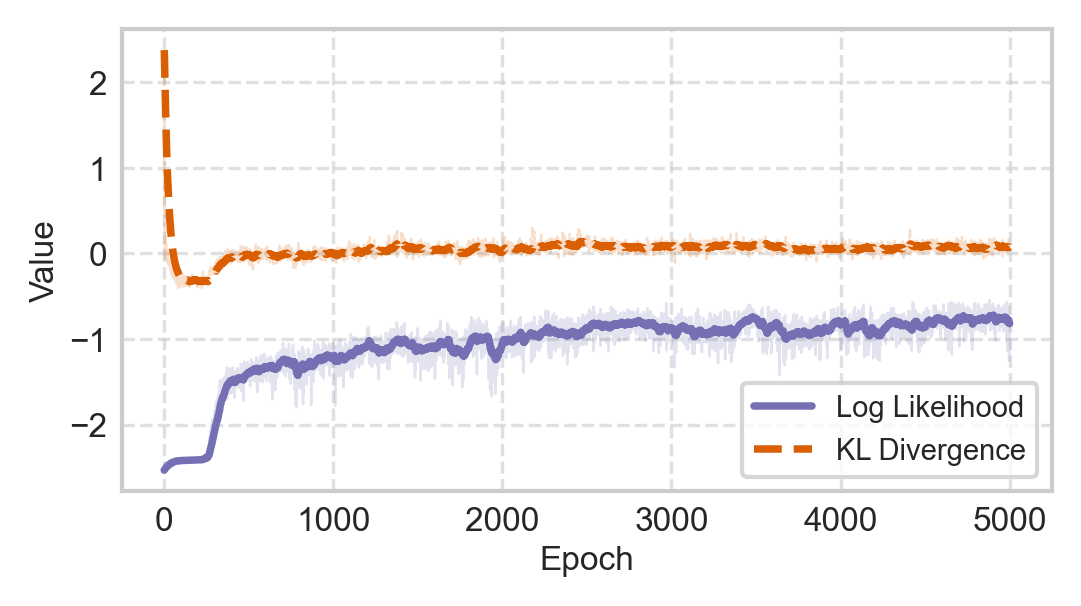}
        \caption{ELBO component dynamics.}
        \label{fig:elbo_dynamic}
    \end{subfigure}
    
    \caption{Left: Action prediction accuracy over training epochs for intermediate actions and final evacuation decisions. 
    Right: Evolution of the action likelihood term and the KL divergence.}
\end{figure}

\noindent\textbf{Action Prediction and ELBO change Analysis.}
We first examine training dynamics and action prediction performance.
As shown in Figure~\ref{fig:elbo_dynamic}, the KL term starts at a high positive value and drops sharply in early training, reflecting rapid alignment between the inference distribution and the belief-transition prior. It then stabilizes around 0, indicating a steady balance between prior regularization and data-driven inference.
Meanwhile, the action log-likelihood $\log p_\theta(a_t \mid \mathbf{b}_t)$ increases steadily, indicating that the inferred beliefs become increasingly predictive of actions.
Together, these trends suggest that action-explanatory belief configurations are first identified through inference and subsequently consolidated into stable generative belief dynamics.
Figure~\ref{fig:action_acc} shows action prediction accuracy for both intermediate actions and final evacuation decisions.
The smooth convergence and stable test performance are consistent with the ELBO dynamics, confirming that improved likelihood and stabilized belief learning translate into reliable action prediction.

\textbf{Sensitivity Analysis over $\tau$.}
We study the effect of the temperature parameter $\tau$ in the base unary score defined in Section 3.2:
$\phi^{\text{base}}_{t,i}
= \tau\!\left(
\cos(\mathbf{h}_{t,i}, \mathbf{h}^{\text{Yes}}_{t,i})
- \cos(\mathbf{h}_{t,i}, \mathbf{h}^{\text{No}}_{t,i})
\right).$

The role of $\tau$ is to anchor the semantic orientation of each belief dimension. As shown in Table~\ref{tab:tau_results}, when $\tau$ is reduced or removed, several belief correlations flip sign, indicating instability in semantic alignment. However, the overall pairwise structure and action prediction performance remain relatively stable across different $\tau$ values.
This suggests that $\tau$ mainly serves to fix the direction of the latent belief space, while having limited impact on downstream tasks. In practice, $\tau$ only needs to be sufficiently large to stabilize semantic orientation during early training.

\begin{table}[htbp]
\centering
\footnotesize
\setlength{\tabcolsep}{3pt}
\begin{tabular}{lccc}
\toprule
Methods & $\tau=15$ & $\tau=3$ & $\tau=0$ \\
\midrule
Belief 0 & 0.3311 ($3.27\times10^{-6}$) & 0.3395 ($1.76\times10^{-6}$) & -0.3246 ($5.19\times10^{-6}$) \\
Belief 1 & 0.4099 ($4.73\times10^{-9}$) & 0.3983 ($1.38\times10^{-8}$) & -0.3294 ($3.69\times10^{-6}$) \\
Belief 2 & 0.2952 ($3.72\times10^{-5}$) & -0.2590 ($3.16\times10^{-4}$) & -0.2545 ($4.09\times10^{-4}$) \\
Belief 3 & 0.1848 ($1.09\times10^{-2}$) & -0.1320 ($7.03\times10^{-2}$) & -0.2340 ($1.19\times10^{-3}$) \\
Belief 4 & 0.3757 ($9.98\times10^{-8}$) & 0.2663 ($2.12\times10^{-4}$) & -0.3395 ($1.77\times10^{-6}$) \\
Belief 5 & 0.3824 ($5.64\times10^{-8}$) & 0.3006 ($2.64\times10^{-5}$) & -0.3221 ($6.20\times10^{-6}$) \\
\midrule
Pairwise corr. & 0.6643 ($6.91\times10^{-3}$) & 0.6150 ($1.48\times10^{-2}$) & 0.6929 ($4.19\times10^{-3}$) \\
Action acc     & 0.8443 & 0.8680 & 0.8695 \\
Final action acc & 0.9157 & 0.8470 & 0.8560 \\
\bottomrule
\end{tabular}
\caption{Sensitivity analysis over $\tau$. Each belief entry reports Spearman correlation (p-value in parentheses).}
\label{tab:tau_results}
\end{table}

\textbf{Interpreting the Attention Matrix.}
The self-attention matrix $A_{t,j}[i,k]$ captures how belief $k$ modulates the contribution of belief $i$ to action $j$, i.e., belief–belief interactions conditioned on the target action.

We consider four intermediate actions: preparing and waiting; acting immediately (e.g., evacuating, defending property, or taking shelter); seeking additional information; and no reaction. The final action is binary, indicating whether the resident evacuates or not. 

The belief set consists of six dimensions: potential damage to one’s home, damage to the neighborhood, personal injury, death, injury to others/pets/livestock, and death of others/pets/livestock.

For actions associated with evacuation or immediate response (intermediate $a_0, a_1$ and final evacuation $A_1$), attention is consistently concentrated on beliefs related to personal risk and property damage (e.g., injury, death, and home destruction). This suggests that evacuation decisions are driven by the joint—and often competing—influence of personal safety and asset protection, a trade-off well documented in the Protective Action Decision Model (PADM)~\cite{PADM2012,champ2016}.

In contrast, for inaction (intermediate $a_3$) and non-evacuation ($A_0$), attention shifts toward beliefs about harm to others (e.g., pets or livestock), while personal-risk-related beliefs are relatively down-weighted. This indicates that when perceived self-risk is low, decision-making is less dominated by survival concerns and instead influenced by social and responsibility-related signals. This observation is also consistent with PADM, which highlights the role of social cues and responsibility in shaping protective behavior.

\begin{figure}[H]
    \centering
    \includegraphics[width=1\linewidth]{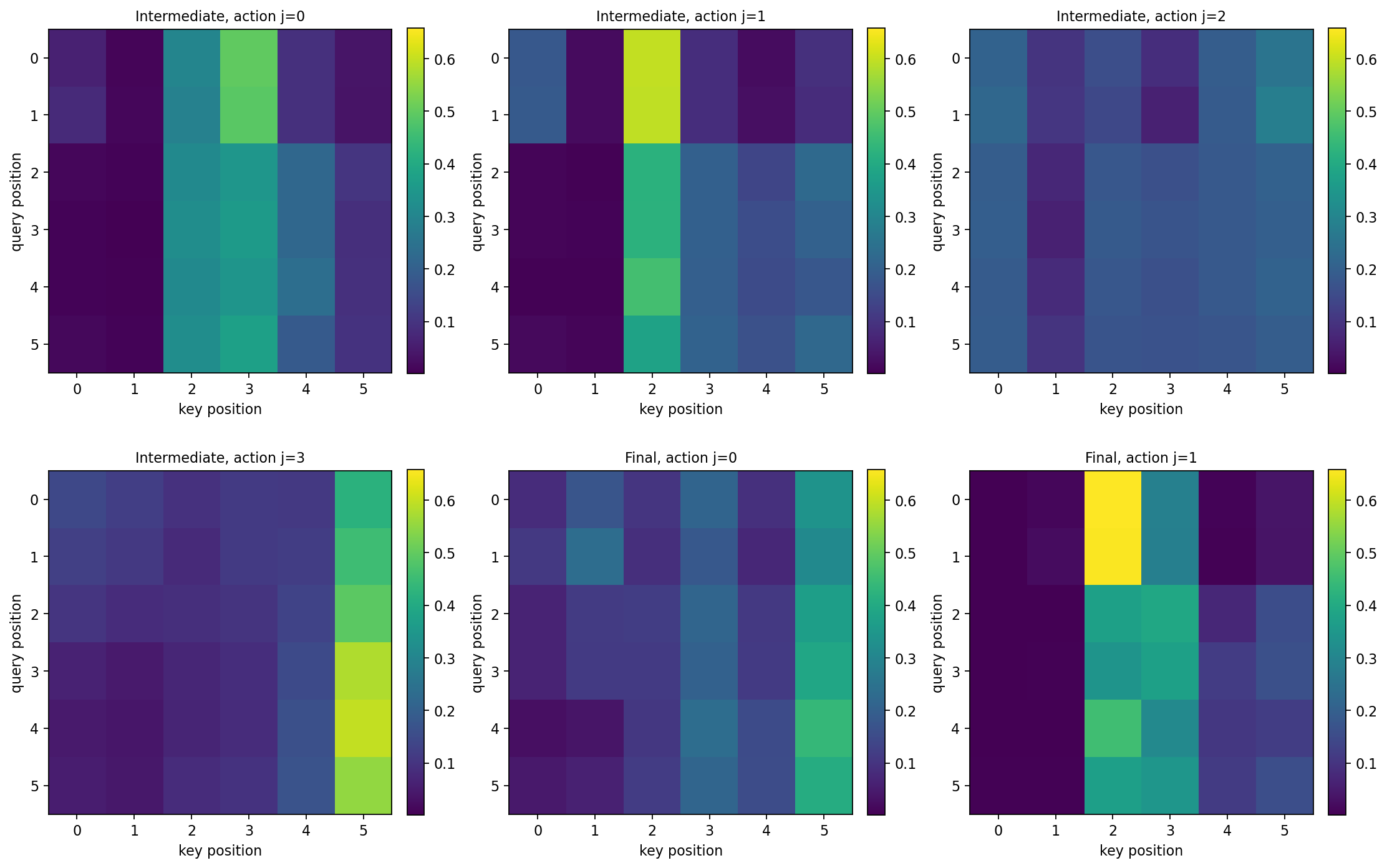}
    \caption{Attention matrix for four intermediate actions and two final actions.}
    \label{fig:att_matrix}
\end{figure}

\clearpage
\section{Beliefs Mediating Role and More Ablation Study}
\label{appx:beliefs_mediating}

\subsection{Beliefs Mediating Role}
To test whether beliefs actually play a mediating role, we performed belief interventions experiments at inference time. Specifically, we keep the observation fixed and manually modify belief logits before the action decoder. We use two types of interventions:
flipping extreme beliefs (e.g., $b<0.2$ flipped to 1, $b>0.8$ flipped to 0)
flipping neutral beliefs (e.g.,$0.4< b<0.5$ flipped to 1, $0.5< b<0.6$ flipped to 0).
We then measure:
flip rate: how often the predicted action changes.
probability change: how much the action probability shifts.
Results in Table ~\ref{tab:belief_intervention} show that changing beliefs leads to clear changes in action predictions. When beliefs are strong (close to 0 or 1), flipping them leads to large changes in action predictions, while flipping weak or uncertain beliefs (around 0.5) has very little effect. This suggests that beliefs function as mediators inside the model, and beliefs the model are confident about have a strong influence on actions. This also corresponds to the core theory in PADM, which strongly emphasize how critical that threat and risk perceptions are for high-stake dynamic decision-making like disaster responses.

\subsection{Additional Ablation on ELBO and Factor Graphs}
It is necessary to disentangle the effect of learning from data and the contribution of the ELBO objective and factor graph structure. To address this, we include additional ablations:
(1) No ELBO, no factor graph: removing pairwise belief interactions so that beliefs are modeled independently, while still treating them as latent variables in the generative process.
(2) No ELBO: removing the variational objective and training the model using only the action likelihood, while still maintaining latent beliefs in the generative process.

As shown in Table ~\ref{tab:ablation_elbo_fg}, removing ELBO leads to a substantial drop in both belief estimation and action prediction. Removing the factor graph results in a drop in pairwise correlation.

\begin{table}[htbp]
\centering
\setlength{\tabcolsep}{2.5pt}
\begin{tabular}{p{6cm}cccc}
\toprule
Belief interventions 
& \shortstack{Inter action\\flip rate} 
& \shortstack{Inter action\\prob. change} 
& \shortstack{Final action\\flip rate} 
& \shortstack{Final action\\prob. change} \\
\midrule

\shortstack[l]{Flipping extreme beliefs\\
($b < 0.2 \rightarrow 1,\ b > 0.8 \rightarrow 0$)}
& 0.6508 & 0.2869 & 0.1633 & 0.2408 \\

\shortstack[l]{Flipping neutral beliefs\\
($0.4 < b < 0.5 \rightarrow 1,\ 0.5 < b < 0.6 \rightarrow 0$)}
& 0.1640 & 0.0550 & 0.0306 & 0.0538 \\

\bottomrule
\end{tabular}
\caption{Belief intervention test. With flip rate and probability change reported for intermediate actions and final actions.}
\label{tab:belief_intervention}
\end{table}

\begin{table}[htbp]
\centering
\small
\setlength{\tabcolsep}{4pt}
\begin{tabular}{lccc}
\toprule
 & Our Model & No ELBO & No ELBO, no factor graph \\
\midrule
Belief 0 & 0.3311 ($3.27 \times 10^{-6}$) & 0.1605 ($2.74 \times 10^{-2}$) & 0.1767 ($1.50 \times 10^{-2}$) \\
Belief 1 & 0.4099 ($4.73 \times 10^{-9}$) & 0.3049 ($2.00 \times 10^{-5}$) & 0.2707 ($1.65 \times 10^{-4}$) \\
Belief 2 & 0.2952 ($3.72 \times 10^{-5}$) & 0.1826 ($1.19 \times 10^{-2}$) & 0.2032 ($5.05 \times 10^{-3}$) \\
Belief 3 & 0.1848 ($1.09 \times 10^{-2}$) & 0.1526 ($3.61 \times 10^{-2}$) & 0.1755 ($1.57 \times 10^{-2}$) \\
Belief 4 & 0.3757 ($9.98 \times 10^{-8}$) & 0.2921 ($4.53 \times 10^{-5}$) & 0.3454 ($1.13 \times 10^{-6}$) \\
Belief 5 & 0.3824 ($5.64 \times 10^{-8}$) & 0.2938 ($4.07 \times 10^{-5}$) & 0.2983 ($3.07 \times 10^{-5}$) \\
\midrule
Pairwise correlation & 0.6643 & 0.5464 & 0.4857 \\
Pairwise p value & $6.91 \times 10^{-3}$ & $3.51 \times 10^{-2}$ & $6.64 \times 10^{-2}$ \\
\midrule
Action acc & 0.8443 & 0.4794 & 0.4802 \\
Final action acc & 0.9157 & 0.7830 & 0.7806 \\
\bottomrule
\end{tabular}
\caption{Ablation on ELBO and factor graph. Belief rows show Spearman correlation (p-value in parentheses); pairwise correlation captures belief relations; action metrics are prediction accuracy.}
\label{tab:ablation_elbo_fg}
\end{table}

\clearpage
\section{Beliefs Analysis}
\label{app:belief_analysis}

We consider six belief dimensions derived from the survey, capturing residents’ perceived wildfire risk at the property, self-safety, and others-safety levels. Specifically, the first two beliefs concern potential damage to one’s home and neighborhood (Table~\ref{tab:belief_definitions}).

\subsection{Belief Update Trajectories}
Throughout the paper, beliefs are explicitly modeled as time-varying latent states and jointly inferred across multiple time steps. Here, we complement this formulation by clustering and visualizing belief trajectories directly from the survey data, providing empirical evidence that such temporal belief dynamics are present in the data and exhibit structured heterogeneity across individuals.

To focus on the trend and shape of belief evolution rather than the absolute score level, we apply within-resident $z$-score normalization to each three-point trajectory. After normalization, each resident’s trajectory has zero mean and unit variance across time, which removes individual-specific baseline and scale differences in subjective belief ratings and ensures that trajectories are compared based on temporal variation rather than absolute magnitude.

Specifically, for resident $i$ (and a fixed belief dimension), let the observed survey trajectory be $\mathbf{x}_i = [x_{i,0}, x_{i,1}, x_{i,2}]$. We compute the resident-specific mean and standard deviation
\begin{equation}
\mu_i = \frac{x_{i,0}+x_{i,1}+x_{i,2}}{3}, \qquad
\sigma_i = \sqrt{\frac{(x_{i,0}-\mu_i)^2+(x_{i,1}-\mu_i)^2+(x_{i,2}-\mu_i)^2}{3}},
\end{equation}
and normalize each time point as, where $\epsilon$ is a small constant for numerical stability.
\begin{equation}
z_{i,t} = \frac{x_{i,t}-\mu_i}{\sigma_i+\epsilon}, \quad t\in\{0,1,2\},
\end{equation}

\begin{figure}[t]
    \centering
    \includegraphics[width=\linewidth]{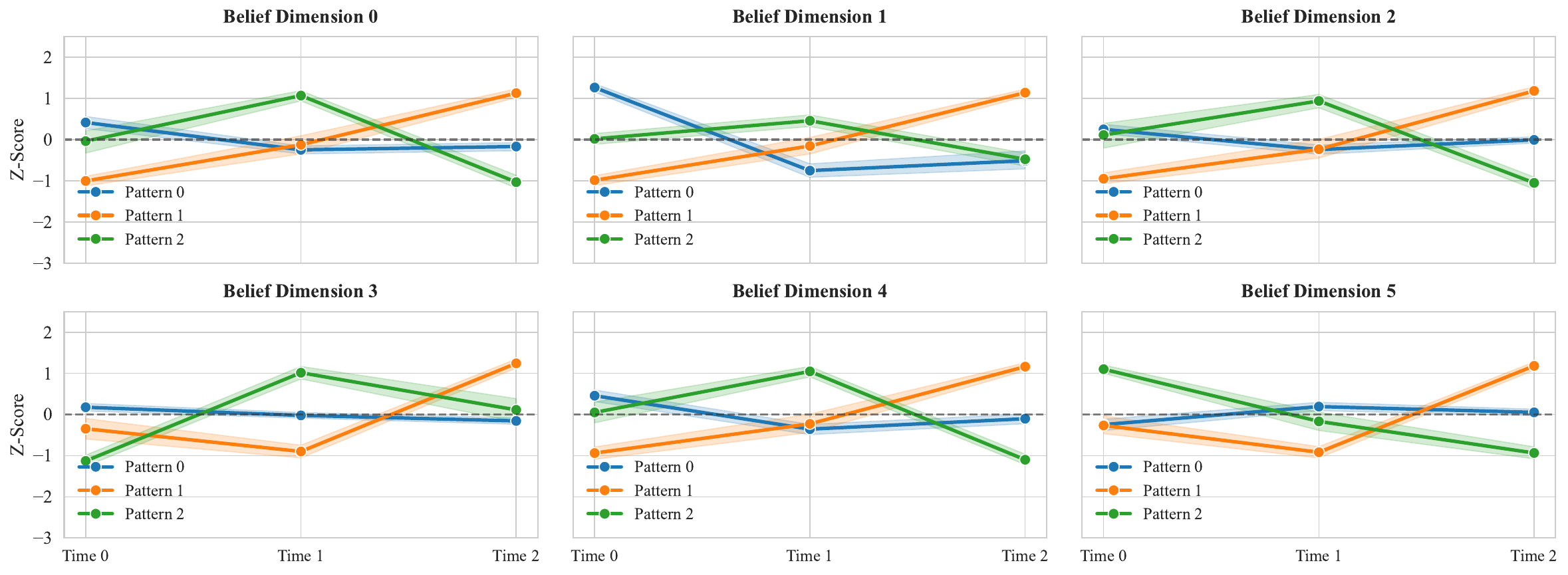}
    \caption{Clustered belief update trajectories derived from survey responses.}
    \label{fig:belief_traj}
\end{figure}

As shown in Figure~\ref{fig:belief_traj}, applying $k$-means clustering with $k=3$ to the normalized trajectories reveals three distinct belief trajectory types: a panic-type pattern characterized by monotonically increasing beliefs over time, a recovery-type pattern exhibiting an initial increase followed by a decline, and a calm-type pattern with relatively stable belief levels across time.

\subsection{Belief Distributions}
\begin{figure}[t]
    \centering
    \includegraphics[width=1\linewidth]{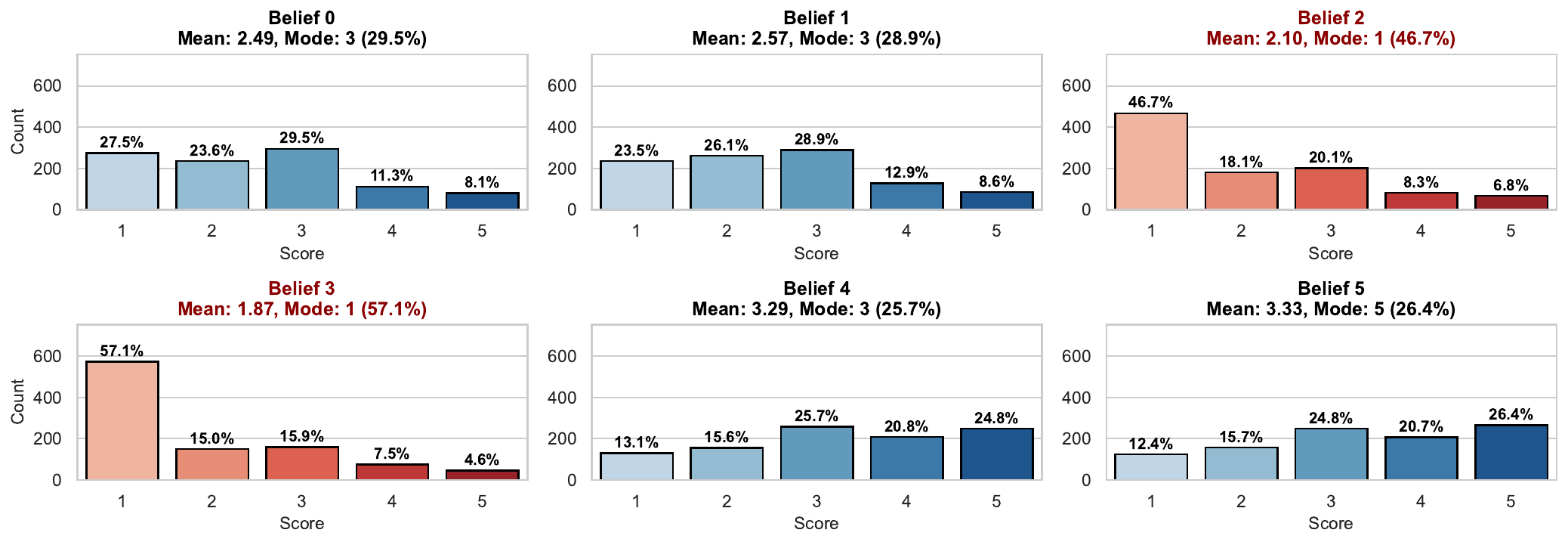}
    \caption{Empirical score distributions of the six survey belief dimensions. }
    \label{fig:belief_distri}
\end{figure}

We examine the empirical score distributions of the six belief dimensions derived from the survey. As shown in Figure~\ref{fig:belief_distri}, while most beliefs exhibit relatively balanced distributions, we observe that Belief~2 (risk to one’s own injury) and Belief~3 (risk to one’s own death) display pronounced long-tailed distributions, with a substantial mass concentrated at low scores.

This pattern is consistent with the well-documented phenomenon of \emph{optimistic bias} in disaster psychology \cite{Kinsey2019, Kinateder2015, Lyu2022}, whereby individuals tend to underestimate the likelihood of negative events occurring to themselves, even when acknowledging risks to property or to others. 
This distributional property provides a plausible explanation for why Belief~2 and Belief~3 exhibit weaker predictive performance compared to other belief dimensions in the main results. The strong skewness and limited variance in these beliefs reduce the effective signal available for modeling and correlation analysis, leading to systematically lower performance relative to beliefs with more balanced score distributions.

\begin{table}[t]
\centering
\caption{Survey belief dimensions and semantic grouping.}
\label{tab:belief_definitions}
\begin{tabular}{llp{9cm}}
\toprule
\textbf{Category} & \textbf{Belief ID} & \textbf{Belief Statement} \\
\midrule
\multirow{2}{*}{Property Risk} 
& B0 & My home would be damaged or destroyed by fire. \\
& B1 & My neighborhood would be damaged or destroyed by fire. \\
\midrule
\multirow{2}{*}{Self Safety Risk} 
& B2 & I might become injured. \\
& B3 & I might die. \\
\midrule
\multirow{2}{*}{Others Safety Risk} 
& B4 & Other people, pets, or livestock might become injured. \\
& B5 & Other people, pets, or livestock might die. \\
\bottomrule
\end{tabular}
\end{table}

\clearpage
\section{Experiments Details}
\label{appx:more_experiments}
We provide additional implementation details, ablations, and analyses in the appendix. All experiments are conducted on NVIDIA A100 GPUs. Training, inference, and evaluation use a single GPU, while larger LLM backbones for embedding extraction are run on 2–3 GPUs.

\noindent\textbf{Why ELBO Is Necessary.}
Across both unary belief prediction and pairwise belief structure learning, our model consistently outperforms all baselines.
These results indicate that the ELBO objective is critical for learning meaningful latent beliefs, as it forces belief variables to jointly explain observations and actions, positioning beliefs as a mediating latent layer rather than post-hoc representations.
Crucially, this alignment is achieved without belief-level supervision, which is essential in real-world settings where reliable belief annotations are unavailable.
In contrast, inference-time prompting or heuristic belief extraction lacks optimization-based constraints tying beliefs to actions, leading to weakly identifiable belief representations that are highly sensitive to model priors.

\noindent\textbf{Ablation Study: Roles of Each Component.}
When removing the pairwise interaction component $\phi_{ij}$, the pairwise Spearman correlation drops.
This gap indicates that inter-belief structure is hard to recover from marginal belief prediction or action supervision alone.
The reason is that $\phi_{ij}$ provides a direct modeling pathway for belief--belief dependencies, allowing the ELBO objective to allocate gradients specifically to pairwise interactions rather than relying solely on indirect signals mediated through actions or marginal belief terms.
Removing temporal modeling degrades trajectory-level metrics.
This occurs because the temporal component enables gradients to propagate across the entire time horizon, allowing beliefs to accumulate evidence and maintain consistency over long-term decision trajectories.
In contrast, the per-belief Spearman correlation is relatively insensitive to the removal of structural or temporal components.
This suggests that individual belief dimensions are primarily learned through the ELBO objective itself, which provides the dominant learning signal for aligning each belief with observed actions.
Taken together, the ablation study confirms a clear division of labor: ELBO-based training determines what beliefs are present, the pairwise component learns how beliefs interact, and the temporal component captures how beliefs evolve over time.
\paragraph{Generality of the Framework.}
The framework is general at both the modeling and problem levels.
At the modeling level, it imposes no restrictions on dimensionality: individual beliefs are represented by unary energies $\phi_i$, belief--belief interactions by pairwise energies $\phi_{ij}$, and actions by unary energy terms with a softmax, allowing belief and action spaces of arbitrary sizes to be learned jointly. For larger belief spaces ($K>6$), approximate inference methods such as mean-field or MCMC can be used in place of exact enumeration; we empirically study these approximations in Appendix~\ref{appx:scalability}.

At the problem level, the framework applies to human decision-making settings that follow a common psychological reasoning process, in which observations are interpreted into internal beliefs that in turn guide actions.
Structure learning and an ELBO objective constrain this reasoning to ensure coherent belief propagation and behavioral grounding.
Beyond human cognition, the same formulation extends to general reasoning settings in which observations give rise to latent variables that interact to support downstream conclusions.
For example, in medical reasoning, clinical notes and patient reports induce latent variables representing underlying conditions, whose interactions determine diagnostic outcomes; in legal reasoning, evidence descriptions give rise to latent variables such as intent that jointly support legal judgments.

\paragraph{Implications for LLM-Based Reasoning}
Our framework positions LLMs within a structured Theory-of-Mind reasoning process guided by dynamic belief graphs, offering a principled approach for building autonomous agents.
In terms of alignment, existing methods either rely on implicit optimization signals such as RLHF~\cite{ouyang2022}, which lack interpretability, or on rule-based prompting, which lacks flexibility~\cite{zhao2021calibrate,valmeekam2023on}. Our approach provides a middle ground by explicitly modeling latent belief graphs, enabling interpretable reasoning, auditability, and targeted intervention through belief-level modifications.
For long-horizon reasoning, LLMs often suffer from context drift as irrelevant information accumulates~\cite{liu-etal-2024, shi2023l}. Our temporal belief prior $p_\theta(\mathbf{b}_t \mid \mathbf{b}_{t-1}, o_t)$ serves as a compact and structured memory, maintaining temporal consistency by selectively updating and preserving beliefs over time.
Finally, unlike implicit personalization methods based on retrieval or fine-tuning~\cite{zhang2025personalizat}, our framework represents individuals as explicit, evolving belief graphs. This enables transparent and controllable personalization through direct manipulation of belief states and their interactions.

\clearpage
\section{Selected Questions in Survey Data}
\label{appx:select_q}

\begin{table}[h!]
    \centering
    \caption{Example of the Selected Survey Questions}
    \label{tab:threat_questions}
    \renewcommand{\arraystretch}{1.1}
    \begin{tabularx}{\linewidth}{@{} p{0.15\linewidth} X @{}}
        \toprule
        \textbf{Category} & \textbf{Survey Question} \\
        \midrule

        \textbf{Observations} &
        \textbf{(1)} Before you decided to evacuate (or stay), did one or more emergency officials let you know that your area was under an evacuation warning, pre-evacuation notice, or an evacuation order?\\
        & \textbf{(2)} Before you decided to evacuate (or stay), did someone you know tell you to evacuate or that an evacuation warning or order was issued for your area?\\
        & \textbf{(3)} Before you decided to evacuate (or stay), did you see, hear, or feel flames or embers in your immediate vicinity (that is, your neighborhood)?\\
        \midrule

        \textbf{Beliefs} &
        \textbf{(1)} After first finding out from the emergency official(s), how much did you believe each of the following statements on a scale of 1 to 5?\\
        & \textbf{(2)} After first finding out from someone you know, how much did you agree with each of the following statements on a scale of 1 to 5?\\
        & \textbf{(3)} After you first observed fire cues, how much did you believe each of the following statements on a scale of 1 to 5?\\
        & \textbf{(A)} My home would be damaged or destroyed by fire. \\
        & \textbf{(B)} My neighborhood would be damaged or destroyed by fire. \\
        & \textbf{(C)} I might become injured. \textbf{(D)} I might die. \\
        & \textbf{(E)} Other people, pets or livestock might become injured. \\
        & \textbf{(F)} Other people, pets or livestock might die. \\
        \midrule

        \textbf{Actions} &
        \textbf{(1)} What was your immediate reaction when the emergency official(s) first let you know?\\
        & \textbf{(2)} What was your immediate reaction when someone you know first told you?\\
        & \textbf{(3)} What was your immediate reaction to observing fire cues?\\
        & \textbf{(A)} No reaction; I continued my activities. \textbf{(B)} I tried to find more information. \\
        & \textbf{(C)} I started preparing to act, and then waited. 
         \textbf{(D)} No reaction; I continued my activities. \\
        \midrule

        \textbf{Past Experiences} &
        \textbf{(1)} Before the Kincade fire, how many times in the past 10 years did you evacuate because of a wildfire?\\
        & \textbf{(2)} How long had you lived at that residence?\\
        & \textbf{(3)} Before the Kincade fire, did you know that wildfires could be a problem in your community?\\
        & \textbf{(4)} Before the Kincade fire, had you or others taken any measures to protect your residence from wildfires?\\
        \midrule

        \multirow{5}{=}{\textbf{Demographic Info}}&
        \textbf{(1)} How old are you?\\
        & \textbf{(2)} What is the highest level of education you have completed?\\
        & \textbf{(3)} Which of the following categories best describes your employment status at the time the Kincade fire started?\\
        & \textbf{(4)} How many children under 13 years old lived in your household at the time the Kincade fire started?\\
        & \textbf{(5)} How many adults (18 years old to 64 years old) lived in your household at the time the Kincade fire started?\\
        \midrule

        \textbf{T-1 Beliefs} &
        \textbf{(1)} Before the Kincade fire, how would you have described the possibility that a wildfire would threaten your property, on a scale from 1 to 5?\\
        & \textbf{(2)} Please rate the extent to which you agreed or disagreed with each statement (before or at the time of the Kincade fire).\\
        & \textbf{(A)} I believed that my home was well-built and safe for a wildfire.\\
        & \textbf{(B)} I believed that staying at home was safer than moving to a nearby building or shelter.\\

        \bottomrule
    \end{tabularx}
\end{table}

\clearpage
\section{Prompts}
\label{appx:prompts}
{
\begin{longtable}{@{} p{0.2\linewidth} | p{0.76\linewidth} @{}}
    \caption{List of LLM Prompts used in the experiments.} \label{tab:prompts_list} \\
    
    \toprule
    \textbf{Task Type} & \textbf{Prompt Template Content} \\
    \midrule
    \endfirsthead

    \multicolumn{2}{c}{{\bfseries \tablename\ \thetable{} -- continued from previous page}} \\
    \toprule
    \textbf{Task Type} & \textbf{Prompt Template Content} \\
    \midrule
    \endhead

    \midrule
    \multicolumn{2}{r}{{Continued on next page}} \\
    \bottomrule
    \endfoot

    \bottomrule
    \endlastfoot

    \textbf{Probability of \newline beliefs given \newline observations} & 
    \textbf{Context:} You are a resident in a wildfire evacuation scenario. Based on your history provided above, analyze the current situation. \newline
    \textbf{Hypothetical Scenario:} \newline
    \textbf{1.} Your Past Observation: \texttt{\{past\_observation\}} \newline
    \textbf{2.} Your Current Observation: \texttt{\{observation\}} \newline
    \textbf{3.} Your Previous Belief State: \texttt{\{last\_belief\}} \newline
    \textbf{Hypothesis to Evaluate:} Target Belief: \texttt{\{belief\}} \newline
    \textbf{Task:} Evaluate if you are likely to hold the "Target Belief" given the observation and their previous belief state. Answer with only the letter (A or B). \newline
    \textbf{Options:} (A) Likely \quad (B) Unlikely \\
    \midrule

    \textbf{Probability of \newline actions given \newline beliefs} & 
    \textbf{Context:} You are a resident in a wildfire evacuation scenario. Based on your history provided above, analyze the current situation. \newline
    \textbf{Hypothetical Scenario:} \newline
    \textbf{1.} You hold this belief: \texttt{\{belief\}} \newline
    \textbf{2.} You are evaluating the Action: \texttt{\{action\}} \newline
    \textbf{Task:} Analyze if this specific belief \textbf{encourages} or \textbf{discourages} you to take the target action. Considering your background, is the target action likely? Answer with only the letter (A or B). \newline
    \textbf{Options:} (A) Likely \quad (B) Unlikely \\
    \midrule

    \textbf{Describe states \newline after actions} & 
    \textbf{Context:} You are a resident in a wildfire evacuation scenario. Based on your history provided above, analyze the current situation. \newline
    \textbf{Current Situation:} \texttt{\{observation\}} \newline
    \textbf{Your Decision:} \texttt{\{action\}} \newline
    \textbf{Task:} Describe the IMMEDIATE consequences of this action from your first-person perspective. Focus on sensory details and safety status change. \newline
    \textbf{Constraints:} Keep it brief (1-3 sentences); Be objective; Avoid speculation. \\
    \midrule

    \textbf{Belief \newline Relations} & 
    \textbf{Context:} You are a resident in a wildfire evacuation scenario. Analyze the relationship between two beliefs held by yourself. \newline
    \textbf{Input:} Belief 1: \texttt{\{belief 1\}} \quad Belief 2: \texttt{\{belief 2\}} \newline
    \textbf{Task:} Output only one number: a value between 0 and 1 indicates they weaken each other; a value greater than 1 indicates they enhance each other. Do not explain. \\
    \midrule

    \textbf{Infer belief \newline given action \newline and observation} & 
    \textbf{Context:} You are a resident in a wildfire evacuation scenario. Based on your history provided above, analyze the current situation. \newline
    \textbf{Hypothetical Scenario:} \newline
    \textbf{1.} Resident's Current Observation: \texttt{\{observation\}} \newline
    \textbf{2.} The resident has made this action: \texttt{\{action\}} \newline
    \textbf{Hypothesis to Evaluate:} Target Belief: \texttt{\{belief\}} \newline
    \textbf{Task:} Evaluate if the resident likely holds the "Target Belief" given the observation and the action they made. Answer with only the letter (A or B). \newline
    \textbf{Options:} (A) Likely \quad (B) Unlikely \\
\end{longtable}
}

\newpage

\end{document}